\documentclass[10pt,twocolumn,letterpaper]{article}

\usepackage[pagenumbers]{cvpr} 

\usepackage{graphicx}
\usepackage{amsmath}
\usepackage{amssymb}
\usepackage{multirow}
\usepackage{booktabs}
\usepackage{array}
\usepackage{subcaption}
\usepackage{xcolor}
\usepackage{colortbl} 
\usepackage{graphics}

\definecolor{cvprblue}{rgb}{0.21,0.49,0.74}
\usepackage[pagebackref,breaklinks,colorlinks,citecolor=cvprblue]{hyperref}
\graphicspath{{figures/} }

\title{Joker: Conditional 3D Head Synthesis with Extreme Facial Expressions}

\author{Malte Prinzler$^{1,2}$~~~~Egor Zakharov$^{3}$~~~~Vanessa Sklyarova$^{1,2}$~~~~Berna Kabadayi$^{1}$~~~~Justus Thies$^{1,4}$
\vspace{0.2cm}
\\
$^1$Max Planck Institute for Intelligent Systems, Tübingen\\
$^2$Max Planck ETH Center for Intelligent Systems\\
$^3$ETH Zürich~~~~
$^4$Technical University of Darmstadt
}

\renewcommand{\paragraph}[1]{\smallskip\noindent\textbf{#1}}

\begin{document}
\twocolumn[{%
    \renewcommand\twocolumn[1][]{#1}%
    \vspace{-0.3cm}
    \maketitle
    \vspace{-0.4cm}
    \begin{center}
        \centering
        \captionsetup{type=figure}
        \vspace{-0.5cm}
        \def\svgwidth{\textwidth}
\begingroup%
  \makeatletter%
  \providecommand\color[2][]{%
    \errmessage{(Inkscape) Color is used for the text in Inkscape, but the package 'color.sty' is not loaded}%
    \renewcommand\color[2][]{}%
  }%
  \providecommand\transparent[1]{%
    \errmessage{(Inkscape) Transparency is used (non-zero) for the text in Inkscape, but the package 'transparent.sty' is not loaded}%
    \renewcommand\transparent[1]{}%
  }%
  \providecommand\rotatebox[2]{#2}%
  \newcommand*\fsize{\dimexpr\f@size pt\relax}%
  \newcommand*\lineheight[1]{\fontsize{\fsize}{#1\fsize}\selectfont}%
  \ifx\svgwidth\undefined%
    \setlength{\unitlength}{1202.39996338bp}%
    \ifx\svgscale\undefined%
      \relax%
    \else%
      \setlength{\unitlength}{\unitlength * \real{\svgscale}}%
    \fi%
  \else%
    \setlength{\unitlength}{\svgwidth}%
  \fi%
  \global\let\svgwidth\undefined%
  \global\let\svgscale\undefined%
  \makeatother%
  \begin{picture}(1,0.44011976)%
    \lineheight{1}%
    \setlength\tabcolsep{0pt}%
    \put(0,0){\includegraphics[width=\unitlength,page=1]{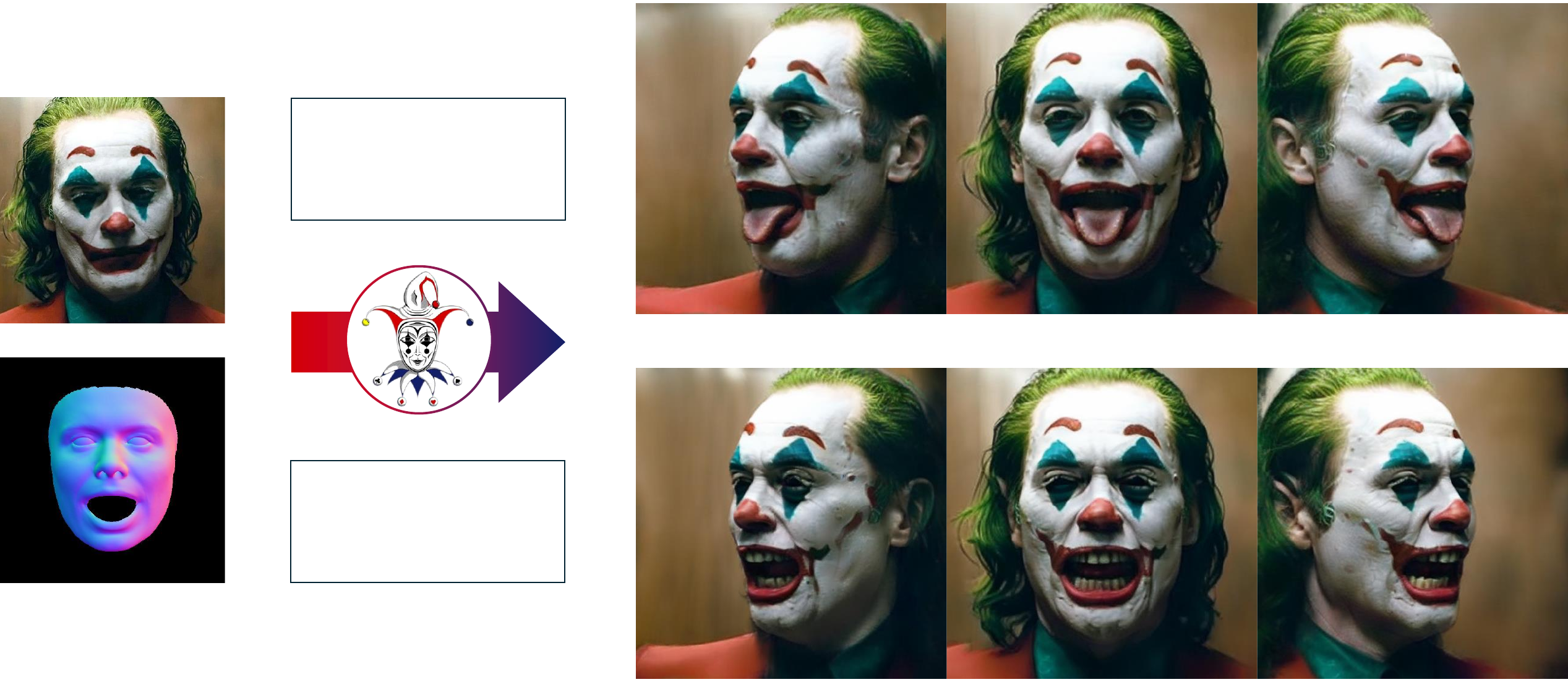}}%
    \put(0.138,0.23468081){\makebox(0,0)[rt]{\lineheight{1.25}\smash{\begin{tabular}[t]{r}\textcolor{white}{\fontsize{8}{8}\textbf{Reference Image}}\end{tabular}}}}%
    \put(0.138,0.069){\makebox(0,0)[rt]{\lineheight{1.25}\smash{\begin{tabular}[t]{r}\textcolor{white}{\fontsize{8}{8}\textbf{3DMM Control}}\end{tabular}}}}%
    \put(0.273,0.347){\makebox(0,0)[t]{\lineheight{1.0}\smash{\begin{tabular}[t]{c}{\fontsize{8}{8}\selectfont\textit{"A man with his}} \end{tabular}}}}%
    \put(0.273,0.325){\makebox(0,0)[t]{\lineheight{1.0}\smash{\begin{tabular}[t]{c}{\fontsize{8}{8}\selectfont\textit{\textbf{tongue sticking out}"}} \end{tabular}}}}%
    \put(0.354,0.30){\makebox(0,0)[rt]{\lineheight{1.25}\smash{\begin{tabular}[t]{r}\textbf{\fontsize{8}{8}\selectfont Text Control}\end{tabular}}}}%
    \put(0.354,0.07){\makebox(0,0)[rt]{\lineheight{1.25}\smash{\begin{tabular}[t]{r}\textbf{\fontsize{8}{8}\selectfont Text Control}\end{tabular}}}}%
    \put(0.273,0.117){\makebox(0,0)[t]{\lineheight{1.0}\smash{\begin{tabular}[t]{c}\fontsize{8}{8}\selectfont\textit{"A man with an} \end{tabular}}}}%
    \put(0.273,0.095){\makebox(0,0)[t]{\lineheight{1.0}\smash{\begin{tabular}[t]{c}\fontsize{8}{8}\selectfont\textit{\textbf{angry expression}"} \end{tabular}}}}%
    \put(0.69926997,0.2107031){\makebox(0,0)[t]{\lineheight{1.25}\smash{\begin{tabular}[t]{c}3D Reconstruction\end{tabular}}}}%
  \end{picture}%
\endgroup%

        \caption{Given a reference image and a target expression defined by a 3DMM and text, Joker generates a 3D model of the subject.}
        \label{fig:teaser}
    \end{center}%
}]

\begin{abstract}
We introduce \emph{Joker}, a new method for the conditional synthesis of 3D human heads with extreme expressions.
Given a single reference image of a person, we synthesize a volumetric human head with the reference's identity and a new expression.
%
%
%
We offer control over the expression via a 3D morphable model (3DMM) and textual inputs.
This multi-modal conditioning signal is essential since 3DMMs alone fail to define subtle emotional changes and extreme expressions, including those involving the mouth cavity and tongue articulation. 
Our method is built upon a 2D diffusion-based prior that generalizes well to out-of-domain samples, such as sculptures, heavy makeup, and paintings while achieving high levels of expressiveness. 
To improve view consistency, we propose a new 3D distillation technique that converts predictions of our 2D prior into a neural radiance field (NeRF).
Both the 2D prior and our distillation technique produce state-of-the-art results, which are confirmed by our extensive evaluations.
Also, to the best of our knowledge, our method is the first to achieve view-consistent extreme tongue articulation. 
\href{https://malteprinzler.github.io/projects/joker/}{Project Page}
\vspace{-0.4cm}
\end{abstract}    
\section{Introduction}
Human head avatars have manifold applications in areas such as AR/VR telepresence~\cite{10.1145/3407662.3407756, tran2024voodooxp, Ma_2021_CVPR}, video games~\cite{Zhu2020ReconstructingNP, Sklyarova2023HAARTG}, and visual effects~\cite{Naruniec2020HighResolutionNF, Li20233DAwareFS}.
To facilitate downstream applications, techniques for creating 2D head avatars from a single image, often controlled by keypoints, parametric models, and other driving modalities, have been widely studied~\cite{zakharov2019few, Wiles2018X2FaceAN, siarohin2019first, drobyshev2022megaportraits, drobyshev2024emoportraits, wang2021one, hong2022depth, xie2024xportrait, ye2024real3dportrait}.
However, in many cases, a view-consistent 3D model of a human subject is required to enable rendering in fully virtual environments.

To address this problem, multiple techniques for reconstructing a 3D model of the human subject have been developed~\cite{tran2023voodoo, tran2024voodooxp, deng2024portrait4d, deng2024portrait4dv2, Khakhulin2022ROME, trevithick2023real, li2023generalizable}.
Human expressions and appearances, however, have a long-tailed distribution.
Mouth cavities and tongues are examples of the areas that are traditionally difficult to capture and, thus, present a challenge for modern avatar systems.
Large head rotations are further challenging scenarios, as such examples are missing in most of the existing human-centric datasets~\cite{zhu2022celebv, yu2022celebvtext, voxceleb, Karras_2019_CVPR}.
Some approaches~\cite{zhang2024rodinhd, avatarsdk, pinscreenavatar} addressed this problem by training avatar reconstruction methods using synthetic datasets of digital human assets rendered via classical graphics pipelines.
However, their photorealism and expressiveness still remain subpar. 
Another group of works~\cite{giebenhain2023nphm, Giebenhain2023MonoNPHMDH, Ma_2021_CVPR, Lombardi2018DeepAM} extended linear parametric head models~\cite{Blanz1999AMM, bfm09, FLAME:SiggraphAsia2017} with non-linear neural components trained from multi-view datasets~\cite{kirschstein2023nersemble, wuu2022multiface}.
However, collecting such data is expensive, and these approaches can not create avatars with controllable tongues or from a single image.
To address these challenges, we propose a novel approach with multi-modal control for extreme expression synthesis. 
Our method follows an existing line of works on human-centric image synthesis~\cite{gu2023diffportrait3d, guo2023animatediff, xie2024xportrait} and fine-tunes a pre-trained Stable Diffusion~\cite{rombach2021highresolution} model paired with a ControlNet~\cite{zhang2023adding}.
We then introduce multi-modal driving inputs to achieve robust and realistic novel-view synthesis of rare and extreme expressions.
Our conditioning signal combines the parameters of a 3DMM with a textual prompt.
We found that text prompts greatly supplement the control with 3DMM parameters by resolving ambiguities w.r.t. subtle emotional changes and tongue articulations. 

View consistency is achieved by optimizing a 3D NeRF~\cite{mildenhall2020nerf} from the predictions of our diffusion prior using a novel distillation procedure.
Existing 3D distillation approaches \cite{poole2022dreamfusion, wang2023prolificdreamer, wang2023imagedream, shi2023MVDream, wu2023reconfusion, gao2024cat3d, zheng2024unified} exploit 2D diffusion models to predict pseudo-ground-truth target images from noised renders of the current 3D representation. 
Most of these methods update these target images for every optimization step of the 3D representation -- i.e., utilize \emph{dynamic targets}.
In a recent concurrent work~\cite{gao2024cat3d}, it was shown that improved performance can be achieved by generating all pseudo ground truth images only once and then optimizing the 3D representation against the generated \emph{static targets}. 

We found that neither dynamic nor static target-based approaches yield optimal results for novel-expression 3D distillation. 
Instead, we propose a new distillation procedure based on \emph{progressively updated targets}.
For each time step of a standard DDIM~\cite{song2022denoising} denoising schedule, we use a diffusion-based prior to predict all target images from noised renderings of the 3D representation.
We then optimize the 3D representation for several iterations against these target images.
Notably, we found it highly beneficial to transition from the dynamically updated target images to static target images at some point during the optimization process.
Thus, our proposed progressive distillation method consists of two stages: i) optimization based on dynamically updated targets and ii) optimization based on static targets.
We demonstrate that this procedure converges more stably than dynamic-target approaches and is more robust to multi-view inconsistencies than static-target approaches.
Ultimately, this yields 3D reconstructions of extreme expressions with high visual fidelity. 

To evaluate our method, we collected new benchmark samples that contain extreme expressions in both studio-capture and in-the-wild environments. 
Thus, we evaluate our method on three datasets: our proposed extreme expression benchmark, CelebV-Text~\cite{yu2022celebvtext}, and NeRSemble~\cite{kirschstein2023nersemble}.
We demonstrate an improved performance compared to existing baselines across all these benchmarks. 
To train our method, we have also collected new metadata for the abovementioned datasets, such as textual descriptions of the facial expressions and 3DMM fittings.

\medskip
\noindent
In summary, we contribute:
\begin{itemize}
    \item a 2D diffusion model for single-shot extreme expression synthesis with control through text prompts and 3DMM parameters,
    \item a 3D distillation approach exploiting progressively updated optimization targets to generate photorealistic 3D reconstructions with extreme expressions,
    \item a new benchmark and metadata for existing training datasets tailored for extreme expression synthesis.
\end{itemize}
We plan to make our full codebase, validation dataset, and training dataset metadata publicly available.

\begin{figure*}[t]
    \centering
    \includegraphics[width=\textwidth]{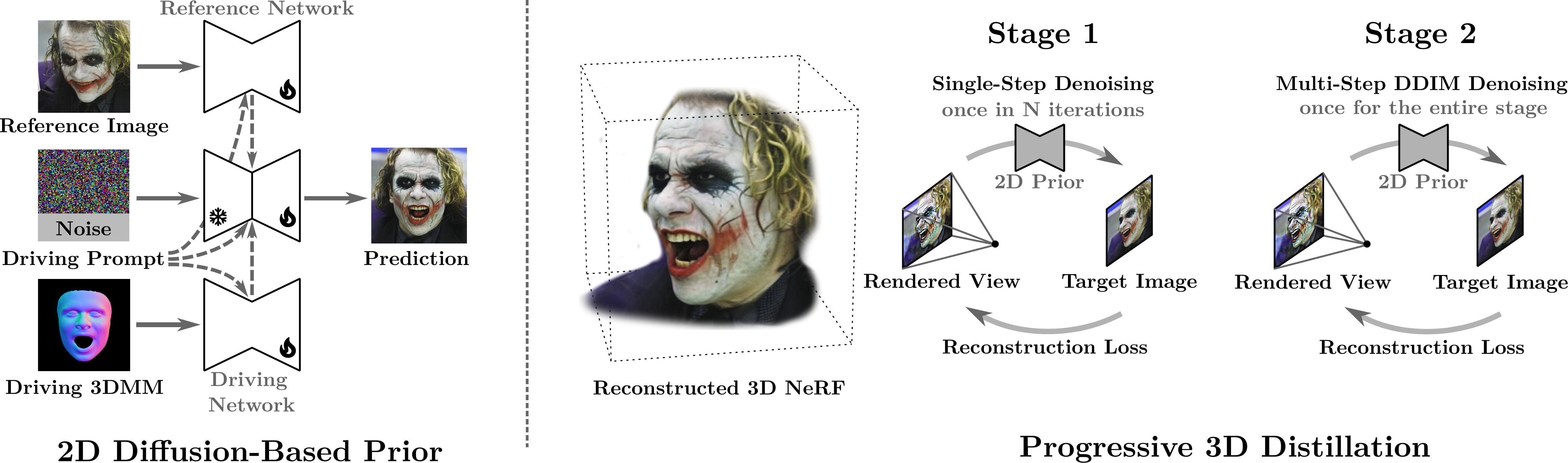}
    \caption{\textbf{Method Overview}. We train a 2D diffusion-based prior for novel pose and expression synthesis from a single reference image. It is controlled through text prompts and 3DMM parameters. We leverage this 2D prior to optimize a Neural Radiance Field (NeRF) \cite{mildenhall2020nerf} with a novel two-stage distillation procedure. During Stage 1, the NeRF is optimized against single-step-denoised predictions of the 2D prior that are recalculated every $N$ optimization iterations. In Stage 2, the target images are calculated once in a multi-step denoising process and kept fixed during the NeRF optimization.}
    \label{fig:pipeline}
\end{figure*}

\section{Related work}

We aim for controllable 3D head synthesis, given a reference image and a target expression.
This target expression is defined in terms of parameters of a 3DMM~\cite{FLAME:SiggraphAsia2017,bfm09} with additional textual control to allow the synthesis of expressions and appearances that fall outside the space of the 3DMM.
Our approach is based on a learned conditional 2D prior which is used for consistent distillation of the 3D head model.

\paragraph{Conditional 2D Head Synthesis.}
Modern 2D head synthesis methods rely on human-centric image- and video-based datasets to train generative models capable of directly synthesizing images with novel expressions from a single portrait image.
Most of these methods utilize GANs~\cite{goodfellow2020generative} or LDMs~\cite{rombach2021highresolution} to enable high realism of the predictions and generalization capabilities of the trained models.
GAN-based reenactment methods~\cite{zakharov2019few, zakharov2020fast, siarohin2019first, siarohin2021motion, doukas2020headgan, drobyshev2022megaportraits, burkov2020neural, drobyshev2024emoportraits, wang2021one} directly predict an output image given a driving signal and a source image.
While achieving high visual quality and expressiveness for frontal-facing images, these methods typically suffer from mode collapse, resulting in low generalization capabilities for extreme expressions and head poses.

Diffusion-based approaches~\cite{xie2024xportrait, chang2024magicpose} resolve some of these limitations by relying on models that were pre-trained on large-scale data, such as Stable Diffusion~\cite{rombach2021highresolution}.
These models can be adapted for human reenactment using a separately trained control network~\cite{zhang2023adding}.
Such an approach achieves a substantially higher degree of generalization than the GAN-based methods trained from scratch using human-centric image and video data.
However, its limitations include a substantial change in the identity of the outputs and a lack of explicit viewpoint control~\cite{xie2024xportrait, chang2024magicpose}.
Moreover, these methods use driving modalities such as keypoints~\cite{chang2024magicpose} or reenacted images produced by a pre-trained GAN-based network~\cite{xie2024xportrait}, which have a limited expressiveness.

\paragraph{Conditional 3D Head Synthesis.}
3D-aware head synthesis methods largely address the challenge of view consistency.
These methods can also be trained using GAN-based training procedures.
Typically, they combine a reconstructed 3D human head model with neural rendering to introduce a high degree of view-consistency~\cite{Khakhulin2022ROME, tran2023voodoo, deng2024portrait4d, deng2024portrait4dv2, tran2024voodooxp, li2023generalizable}.
However, these methods have substantial limitations that include a lack of expressiveness and low quality of rendered images.
Moreover, they still lack view consistency, especially in high-frequency features, since only the coarse head shape is reconstructed explicitly while super-resolution modules hallucinate the remaining details.
These problems are addressed by a growing group of methods that modify pre-trained diffusion models to produce view-consistent renders via viewpoint conditioning~\cite{liu2023zero1to3, wu2023reconfusion, gao2024cat3d}.  
They have been further adapted to human avatar synthesis and can be trained to include explicit pose control~\cite{ho2024sith, anifacediff, bounareli2024diffusionact}.
However, since the views are still predicted in image space directly, the 3D consistency of these methods is subpar.
While some approaches~\cite{gu2023diffportrait3d, xie2024xportrait} attempted to resolve this issue with multi-view-aware denoising techniques, the improvements still remain limited.

An alternative approach is the distillation of pre-trained diffusion models into 3D representations using score distillation sampling (SDS)~\cite{poole2022dreamfusion, wang2023imagedream, wang2023prolificdreamer, jiang2024jointdreaner, shi2023MVDream}, which was used in several previous works on human avatar synthesis~\cite{huang2024tech, liao2023tada}.
However, these methods either fall short in terms of realism~\cite{poole2022dreamfusion} or are unstable w.r.t. the choice of the base diffusion model, i.e. its denoising scheme and hyperparameters, such as classifier-free guidance scale~\cite{wang2023imagedream, wang2023prolificdreamer}.

Contrary to the existing approaches for novel expression synthesis, our method utilizes a progressive optimization strategy of the underlying neural radiance field (NeRF)~\cite{mildenhall2020nerf}.
First, we utilize dynamically updated targets for supervision via single-step denoising, akin to a classical SDS, which results in a blurry yet consistent reconstruction.
Once we achieve a coarse reconstruction, we use it to produce fixed optimization targets using multi-step denoising, which helps to complement the missing high-frequency details.
Compared to using fixed ground-truth targets throughout the distillation process, as in the concurrent work~\cite{gao2024cat3d}, we achieve a substantially higher degree of view consistency, especially for the mouth cavity and tongue.

\begin{figure*}[t]
    \centering
     \def\svgwidth{\linewidth}
    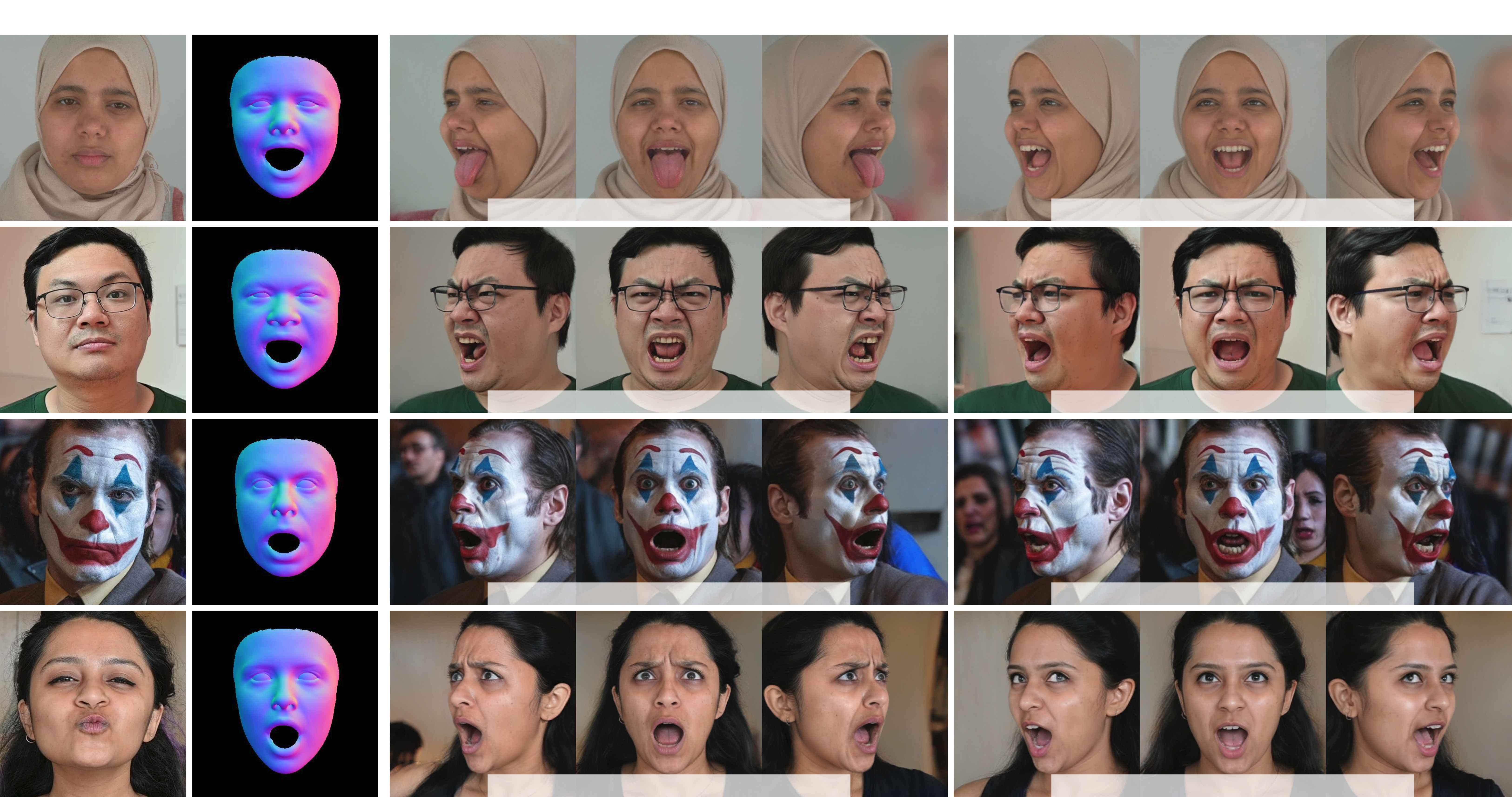
    \caption{\textbf{3DMM- and text-guided 3D reconstruction.} Through text guidance our model resolves ambiguities in the 3DMM control signal, can formulate tongue articulation, and provides fine-grained emotion control. Note that the 3DMM input is kept fixed for both 3D reconstructions of each row and only the text prompt changes.}
    \label{fig:text_control}
    \vspace{-0.3cm}
\end{figure*}

\section{Method}

\label{sec:method}

Our method takes a portrait image as an input and creates a photo-realistic 3D reconstruction with a driving expression specified via parameters of the Basel Face Model \cite{bfm09} and text prompts, see \Cref{fig:pipeline}.
%
%
%
We train a conditional 2D prior that takes an image of a reference person as input and predicts its novel-view renders with novel expressions (\Cref{sec:2dprior}).
%
Exploiting this 2D prior and its predictions, we propose a novel 3D distillation pipeline (\Cref{sec:distillation}) to optimize a view-consistent NeRF.
%

\subsection{2D Prior for Extreme Expression Synthesis}
\label{sec:2dprior}
%


%
Our prior model is based on a Stable Diffusion~\cite{rombach2021highresolution} backbone. 
%
%
%
%
%
%
To convert it into a conditional synthesis model, we follow \cite{xie2024xportrait, chang2023magicdance, anifacediff, gu2023diffportrait3d} and train a separate \emph{reference network} to input the information from the reference image into the denoising network.
%
Specifically, we share keys and values between the self-attention layers of the reference network and its denoising counterpart~\cite{gu2023diffportrait3d}.
%
%
%
We then train a ControlNet~\cite{zhang2023adding}, which we refer to as a \emph{driving network}, to condition our model on the parameters of a 3DMM. 
%
%
%
%
%
Compared to previous methods~\cite{gu2023diffportrait3d, xie2024xportrait} and similar to a concurrent work \cite{anifacediff}, we utilize mesh-based renders of the normal maps to encode the driving head pose and expression.
We have observed that this conditioning style helps to achieve higher view consistency of the results even without view-aware noise or multi-view self-attention techniques~\cite{gu2023diffportrait3d}.
%
%
%

In contrast to existing works \cite{xie2024xportrait, anifacediff, bounareli2024diffusionact}, we also preserve text-based conditioning from the base denoising model and incorporate it into the reference and driving networks.
Text conditioning allows our model to synthesize extreme expressions by supplementing 3DMM-based signals with missing cues, such as tongue movements and emotion-related appearance details.
%
%
%
%
Also, contrary to previous diffusion-based reenactment methods~\cite{xie2024xportrait, anifacediff}, we found it beneficial to fine-tune the decoding part of the denoising network to improve identity preservation.
To train this model, we implemented an iterative training pipeline that utilizes a subset of the in-the-wild CelebV-Text dataset~\cite{yu2022celebvtext} followed by a short fine-tuning phase on the NeRSemble dataset~\cite{kirschstein2023nersemble}.
We supplemented these datasets with new metadata that includes textual annotations of the expressions and 3DMM fittings (\Cref{sec:suppl_datasets}).
%

\subsection{3D Distillation}
\label{sec:distillation}
%
%
Following \cite{wang2023imagedream, shi2023MVDream}, we use a NeRF~\cite{mildenhall2020nerf} formulation to represent our 3D reconstruction. 
%
%
%
During distillation, the NeRF is rendered under several target views and encoded into the latent space of the diffusion model. 
Then, noise is added to these renders, and the 2D diffusion prior denoises the latents and decodes them into the target images $\hat{x}_0$ that the NeRF is optimized against. 

Following previous works~\cite{wu2023reconfusion, shi2023MVDream, wang2023imagedream}, we apply the distillation losses directly in the image domain rather than the latent space.
The optimization objective follows \cite{wu2023reconfusion} and consists of a combination of an L1 distance and a perceptual loss $\mathcal{L}_p$~\cite{lpips} between the rendered images $x$ and targets $\hat{x}_0$:
\begin{equation}
    \mathcal{L}_\text{recon} = \mathbb{E}_{\mathbf{c}} \big[ \| x - \hat{x}_0 \|_1 + \mathcal{L}_p (x, \hat{x}_0) \big],
\end{equation}
where the expectation is taken over the viewing angles.

Our distillation procedure consists of two stages that differ in the way that the target images are updated. 
For Stage 1, the target images are repeatedly updated based on the NeRF renderings to improve their multi-view consistency. 
For Stage 2, the target images are predicted only once and the NeRF is optimized against them until convergence.

\paragraph{Stage 1: Dynamic Target Optimization.}
%
%
%
%
%
%
%
During the first stage, we frequently update the target images using the noised NeRF renderings as input to the distillation prior.
%

We render the NeRF under several target views, apply noise to the renders, and then use the 2D prior to generate the new target images in one denoising step. 
The NeRF is optimized against the target views for $N$ iterations, after which we recalculate the targets and repeat the procedure. 
%
We use 3D-consistent NeRF renderings to update the target images to improve their view consistency. 
The main difference compared to the previous approaches~\cite{wang2023imagedream, shi2023MVDream, wu2023reconfusion} is that we sample the noise levels deterministically following a standard DDIM denoising scheduler with 100 steps, instead of sampling them randomly. 
Furthermore, the NeRF is optimized against the target images for $N$ iterations before updating them whereas most existing methods update the targets at every NeRF update. 
We found that this substantially improves the visual quality and consistency of the reconstruction results (see \Cref{fig:distillation_comparison}).

\begin{figure}[t]
    \centering
  \def\svgwidth{\linewidth}
\begingroup%
  \makeatletter%
  \providecommand\color[2][]{%
    \errmessage{(Inkscape) Color is used for the text in Inkscape, but the package 'color.sty' is not loaded}%
    \renewcommand\color[2][]{}%
  }%
  \providecommand\transparent[1]{%
    \errmessage{(Inkscape) Transparency is used (non-zero) for the text in Inkscape, but the package 'transparent.sty' is not loaded}%
    \renewcommand\transparent[1]{}%
  }%
  \providecommand\rotatebox[2]{#2}%
  \newcommand*\fsize{\dimexpr\f@size pt\relax}%
  \newcommand*\lineheight[1]{\fontsize{\fsize}{#1\fsize}\selectfont}%
  \ifx\svgwidth\undefined%
    \setlength{\unitlength}{1559.99994233bp}%
    \ifx\svgscale\undefined%
      \relax%
    \else%
      \setlength{\unitlength}{\unitlength * \real{\svgscale}}%
    \fi%
  \else%
    \setlength{\unitlength}{\svgwidth}%
  \fi%
  \global\let\svgwidth\undefined%
  \global\let\svgscale\undefined%
  \makeatother%
  \scriptsize
  \begin{picture}(1,1.29828736)%
    \lineheight{1}%
    \setlength\tabcolsep{0pt}%
    \put(0,0){\includegraphics[width=\unitlength,page=1]{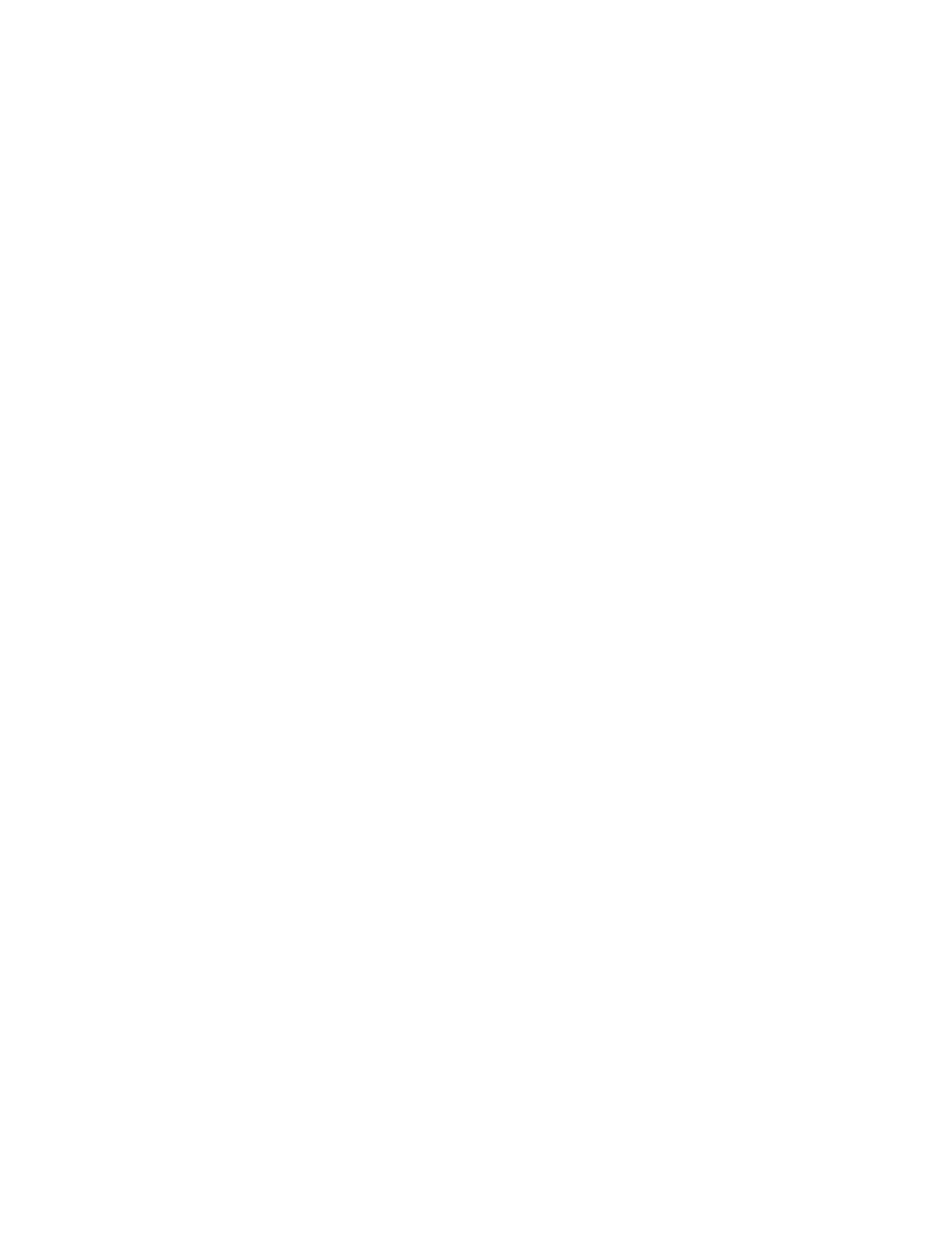}}%
    \put(0.12307692,1.27692322){\makebox(0,0)[t]{\lineheight{1.25}\smash{\begin{tabular}[t]{c}Reference Image\end{tabular}}}}%
    \put(0.63021339,1.27692322){\makebox(0,0)[t]{\lineheight{1.25}\smash{\begin{tabular}[t]{c}3D Reconstruction\end{tabular}}}}%
    \put(0,0){\includegraphics[width=\unitlength,page=2]{ood_v1.pdf}}%
  \end{picture}%
\endgroup%

    \caption{Out-of-distribution 3D reconstruction examples.
    }
    \label{fig:ood}
\end{figure}

\paragraph{Stage 2: Fixed Target Optimization.}
Performing Stage 1 distillation alone tends to drift towards blurry results. 
The reason for this lies in the NeRF optimization against the target images. 
During optimization, the NeRF effectively averages over the inconsistencies in the target images and introduces a low-frequency bias.
This has a significant impact on the distillation procedure. 
After the NeRF optimization, its renderings will be more blurry than the target images.
%
%
%
The lack of high-frequency details in the NeRF renderings is picked up by the 2D prior and propagates into the updated target images. 
Optimizing the NeRF against them leads to even more blurry reconstruction results and introduces a positive feedback loop. 
%
%

%
To effectively solve this phenomenon, we interrupt Stage~1 at an intermediate noise level (after 60 out of 100 denoising steps) and generate the final target images through standard DDIM sampling with the 40 remaining steps.
Afterward, the NeRF is optimized against the final target images until convergence.
While the optimization procedure against the fixed target images has a low-frequency bias as well, we avoid the repeated low-frequency feedback loop of Stage 1 and converge to high-quality results. 
Note that Stage 2 of our distillation procedure is similar to static-target approaches, with the difference that we start the denoising process not from white noise but from view-consistent renderings of a well-converged NeRF.
This largely improves the view consistency of the final target images, while the multi-step denoising procedure generates high-frequency details. 
As a consequence, the predictions of the 2D prior combine high view consistency and image quality which results in superior NeRF optimization results (see \Cref{fig:distillation_comparison}).

\subsection{Implementation Details}
We train the 2D prior starting from pretrained weights of Stable Diffusion v1.5\footnote{\url{huggingface.co/runwayml/stable-diffusion-v1-5}}. 
We use the AdamW optimizer \cite{adamw} with $\text{lr}=10^{-5}, \beta_1=0.9, \beta_2=0.999, \epsilon=10^{-8}, \lambda=10^{-2}$.
Our model is trained on 8 NVIDIA A100 SXM4-80GB with a per-GPU batch size of 10 for 200,000 iterations on CelebV-Text~\cite{yu2022celebvtext} and 30,000 more iterations on an equal mix of CelebV-Text and NerSemble~\cite{kirschstein2023nersemble}.
Please refer to \Cref{sec:suppl_datasets} for dataset and preprocessing details.
For the 3D distillation, we use the threestudio framework~\cite{threestudio2023} and largely follow the configuration of ImageDream~\cite{wang2023imagedream}.
To generate the target images, we render BFM~\cite{bfm09} normal maps on a regular $20\times20$ grid of the frontal hemisphere with an azimuth $\in [-22.5°, 22.5°]$ and elevation $\in [-10°, 10°]$. The images are generated with a classifier-free guidance scale of $19.0$. 
We optimize the NeRF for $N=130$ iterations between each target image update. 
During optimization, we randomly sample 64 patches of size $64\times64$ and gradually increase the resolution of the target images from 64 to 512. 
The optimization takes approximately 3 hours on a single NVIDIA A100 SXM4-80GB GPU. 

\begin{figure*}[t]
    \centering
    \def\svgwidth{\linewidth}
    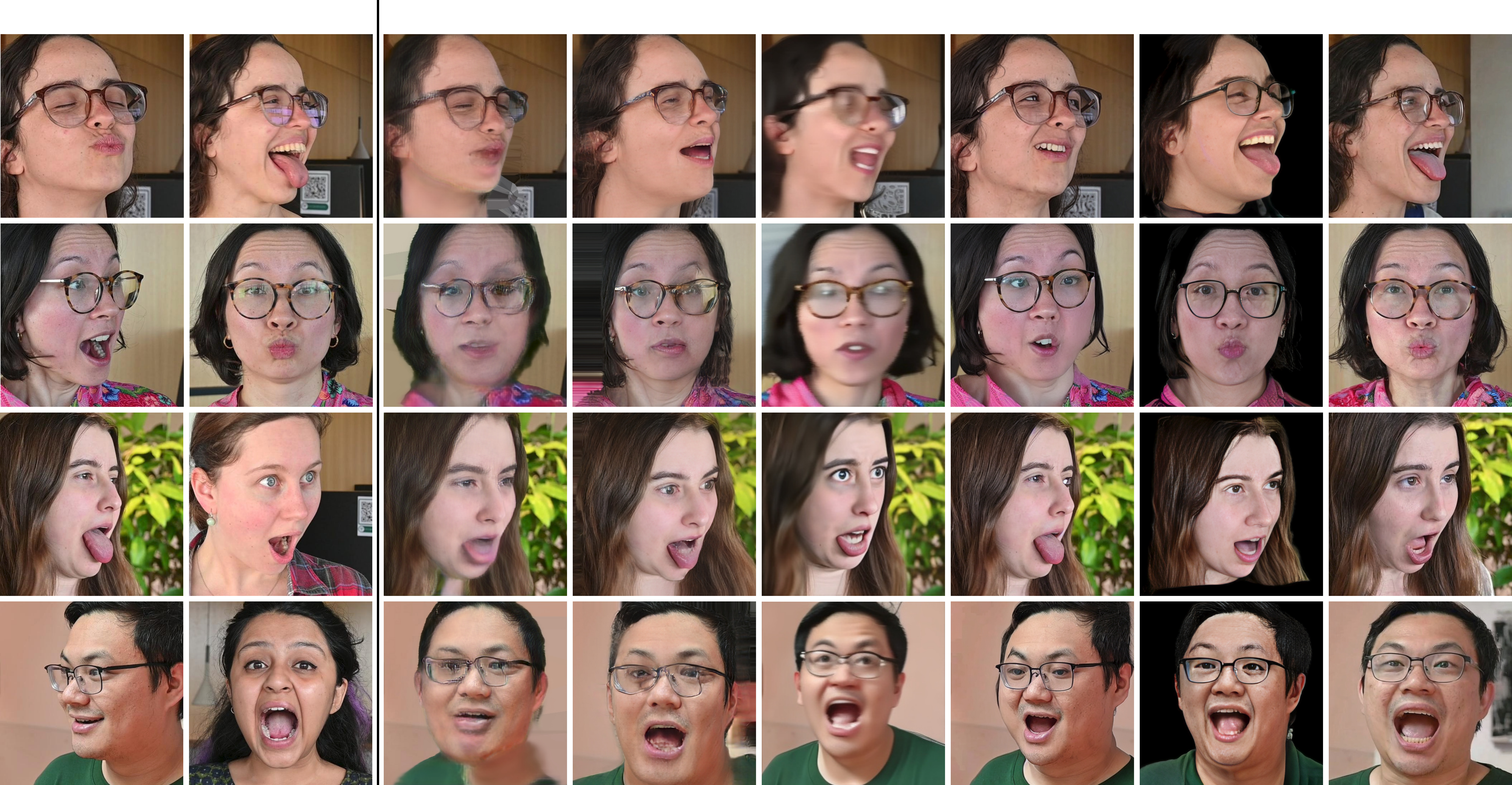
    \caption{Comparison of our \textbf{2D diffusion prior} for self- and cross-reenactment (row 1-2 and 3-4 respectively).}
    \label{fig:2d_prior_comparison}
\end{figure*}

\definecolor{tabfirst}{rgb}{1, 0.7, 0.7} 
\definecolor{tabsecond}{rgb}{1, 0.85, 0.7} 
\definecolor{tabthird}{rgb}{1, 1, 0.7} 

\begin{table*}[t]
    \centering
    \resizebox{\linewidth}{!}{
    \begin{tabular}{l|cccccccc|cccc}
         & \multicolumn{8}{c|}{Self-reenactment} & \multicolumn{4}{c}{Cross-reenactment} \\
         & \multicolumn{1}{c}{PSNR $\uparrow$} & \multicolumn{1}{c}{LPIPS $\downarrow$} & \multicolumn{1}{c}{SSIM $\uparrow$} & \multicolumn{1}{c}{FID $\downarrow$} & \multicolumn{1}{c}{CSIM $\uparrow$} & \multicolumn{1}{c}{AKD $\downarrow$} & \multicolumn{1}{c}{AED $\downarrow$} & \multicolumn{1}{c|}{APD $\downarrow$} & \multicolumn{1}{c}{FID $\downarrow$} & \multicolumn{1}{c}{CSIM $\uparrow$} & \multicolumn{1}{c}{AED $\downarrow$} & \multicolumn{1}{c}{APD $\downarrow$} \\
        \hline

GOHA~\cite{goha}                            &  \cellcolor{tabthird}16.25 & \cellcolor{tabsecond}0.299 &                      0.572 &                      46.33 &                      0.32 & \cellcolor{tabsecond}0.015 &                      0.20 &                      0.42 &                      46.82 &                      0.36 &                      0.26 & \cellcolor{tabsecond}0.51 \\
Real3D-Portrait~\cite{ye2024real3dportrait} & \cellcolor{tabsecond}16.58 &  \cellcolor{tabthird}0.322 & \cellcolor{tabsecond}0.592 &                      31.68 &                      0.46 &  \cellcolor{tabthird}0.021 &                      0.23 &                      0.47 &                      31.54 & \cellcolor{tabsecond}0.57 &                      0.30 &                      0.59 \\
Portrait4D-v2~\cite{deng2024portrait4dv2}   &                      14.13 &                      0.393 &                      0.494 &  \cellcolor{tabthird}19.96 &                      0.49 &                      0.081 &                      0.18 & \cellcolor{tabsecond}0.38 &  \cellcolor{tabthird}20.71 &  \cellcolor{tabthird}0.56 &                      0.25 &  \cellcolor{tabthird}0.52 \\
AniFaceDiff~\cite{anifacediff}              &                      16.24 &                      0.360 &  \cellcolor{tabthird}0.577 &                      29.64 &  \cellcolor{tabthird}0.50 &                      0.034 &  \cellcolor{tabthird}0.17 &  \cellcolor{tabthird}0.41 &                      29.34 &                      0.53 &                      0.25 &  \cellcolor{tabthird}0.52 \\
X-Portrait~\cite{xie2024xportrait}          &                      14.77 &                      0.357 &                      0.493 & \cellcolor{tabsecond}10.66 & \cellcolor{tabsecond}0.61 &                      0.051 &                      0.20 &                      0.68 & \cellcolor{tabsecond}10.79 &  \cellcolor{tabfirst}0.75* &                      0.29 &                      0.81 \\
VOODOO 3D~\cite{tran2023voodoo}             &                      16.11 &                      0.324 &                      0.558 &                      38.63 &                      0.27 &                      0.035 &                      0.19 &                      0.47 &                      39.24 &                      0.30 &  \cellcolor{tabthird}0.24 &                      0.54 \\
VOODOO XP~\cite{tran2024voodooxp}           &                      13.74 &                      0.397 &                      0.483 &                      24.59 &                      0.49 &                      0.075 & \cellcolor{tabsecond}0.15 &                      0.45 &                      24.83 &                      0.43 &  \cellcolor{tabfirst}0.21 &  \cellcolor{tabthird}0.52 \\
\hline
Ours                                         &  \cellcolor{tabfirst}18.63 &  \cellcolor{tabfirst}0.212 &  \cellcolor{tabfirst}0.619 &  \cellcolor{tabfirst}7.57 &  \cellcolor{tabfirst}0.62 &  \cellcolor{tabfirst}0.007 &  \cellcolor{tabfirst}0.11 &  \cellcolor{tabfirst}0.31 &  \cellcolor{tabfirst}8.48 & \cellcolor{tabsecond}0.57 & \cellcolor{tabsecond}0.22 &  \cellcolor{tabfirst}0.49

    \end{tabular}
    }
    \caption{Quantitative comparison of our \textbf{2D diffusion prior} in self- and cross-reenactment scenarios. 
   *: Note that for extreme pose changes, X-Portrait has a tendency to reproduce the reference image without adopting the driving pose, leading to a high identity similarity score (CSIM) yet poor pose accuracy (APD).}
    \label{tab:quant_comparison}
\end{table*}

\section{Experiments}

Below we evaluate both our 2D prior and our 3D distillation technique.
Please refer to \Cref{sec:suppl_additional_experiments} for detailed ablation studies and additional experiments.
To evaluate the synthesis of the extreme facial expressions, the validation sets of CelebV-Text and NeRSemble alone are not sufficient: CelebV-Text only contains moderately extreme expressions, and NeRSemble is restricted to uniform lighting and background scenarios. 
For this reason, we captured the \emph{Joker benchmark} for evaluation of extreme expression synthesis, which will be made publicly available to the research community, see \Cref{sec:suppl_datasets}.
%


\subsection{Evaluation of 2D Prior}

We compare our 2D diffusion prior against the following baselines: 
\emph{GOHA}
~\cite{goha},  
\emph{VOODOO 3D}~\cite{tran2023voodoo}, \emph{VOODOO XP}~\cite{tran2024voodooxp}, \emph{Real3D-Portrait}~\cite{ye2024real3dportrait}, \emph{Portrait4D-v2}~\cite{deng2024portrait4dv2}, \emph{AniFaceDiff}~\cite{anifacediff}, and \emph{X-Portrait}~\cite{xie2024xportrait}.
We refer to \Cref{sec:suppl_baselines} for a detailed discussion of those and their implementation details.
For the baseline results, we used the official code repository of GOHA, VOODOO 3D, Real3D-Portrait, Portrait4D, and X-Portrait; and obtained the results for AniFaceDiff and VOODOO XP from the authors. 

\Cref{fig:2d_prior_comparison} qualitatively compares our 2D diffusion prior with the baselines for the scenario of self- and cross-reenactment. 
We visualize the best-performing baselines and refer the reader to \Cref{fig:2d_prior_comparison_large_self} and \Cref{fig:2d_prior_comparison_large_cross} for further results. 
The comparisons are conducted on the newly captured Joker benchmark, which contains in-the-wild scenes with diverse backgrounds, subject ethnicities, and genders.
We observe a significant qualitative improvement over all baselines.

These results are confirmed by the quantitative comparison, see \Cref{tab:quant_comparison}. 
We evaluate standard metrics such as peak signal-to-noise ratio (PSNR), learned perceptual image patch similarity~\cite{lpips} (LPIPS), structural similarity index measure~\cite{ssim} (SSIM), and Fréchet inception distance~\cite{heusel2017gans} (FID). 
We measure identity preservation by comparing the cosine similarity between the embeddings of a face recognition network \cite{deng2019arcface} for the predicted and ground truth images (CSIM).  
Further, we report the average distance between the extracted keypoints (AKD), expression (AED), and pose (APD) parameters using \cite{deep3dfacerecon}.
For the cross-reenactment scenario where no ground truth data is available, we evaluate the CSIM score between the reference image and the prediction and the AED and APD scores between the driving image and prediction. 

Since VOODOO 3D, VOODOO XP, and GOHA do not synthesize the background, we mask out the backgrounds of the other methods using MODNet~\cite{modnet} before evaluation. 
The quantitative comparison is conducted on 10,000 images for self- and cross-reenactment respectively, evenly sampled from the validation sets of CelebV-Text and NeRSemble, our recordings in a studio environment with uniform lighting and background, and our recordings with in-the-wild backgrounds and lighting. 
\Cref{tab:quant_comparison} shows a consistent improvement over all existing methods. 
Note that VOODOO XP, VOODOO 3D, Real3D-Portrait, GOHA, and Portrait4D are single-step methods that enable real-time inference, while AniFaceDiff, X-Portrait, and our method are diffusion models that require several denoising steps. 
%
Further, X-Portrait was trained on video sequences and with a different crop size than our evaluation samples. However, at the time of writing this paper, no code was available to retrain the model.

\begin{figure}[t]
    \centering
    \def\svgwidth{\linewidth}
    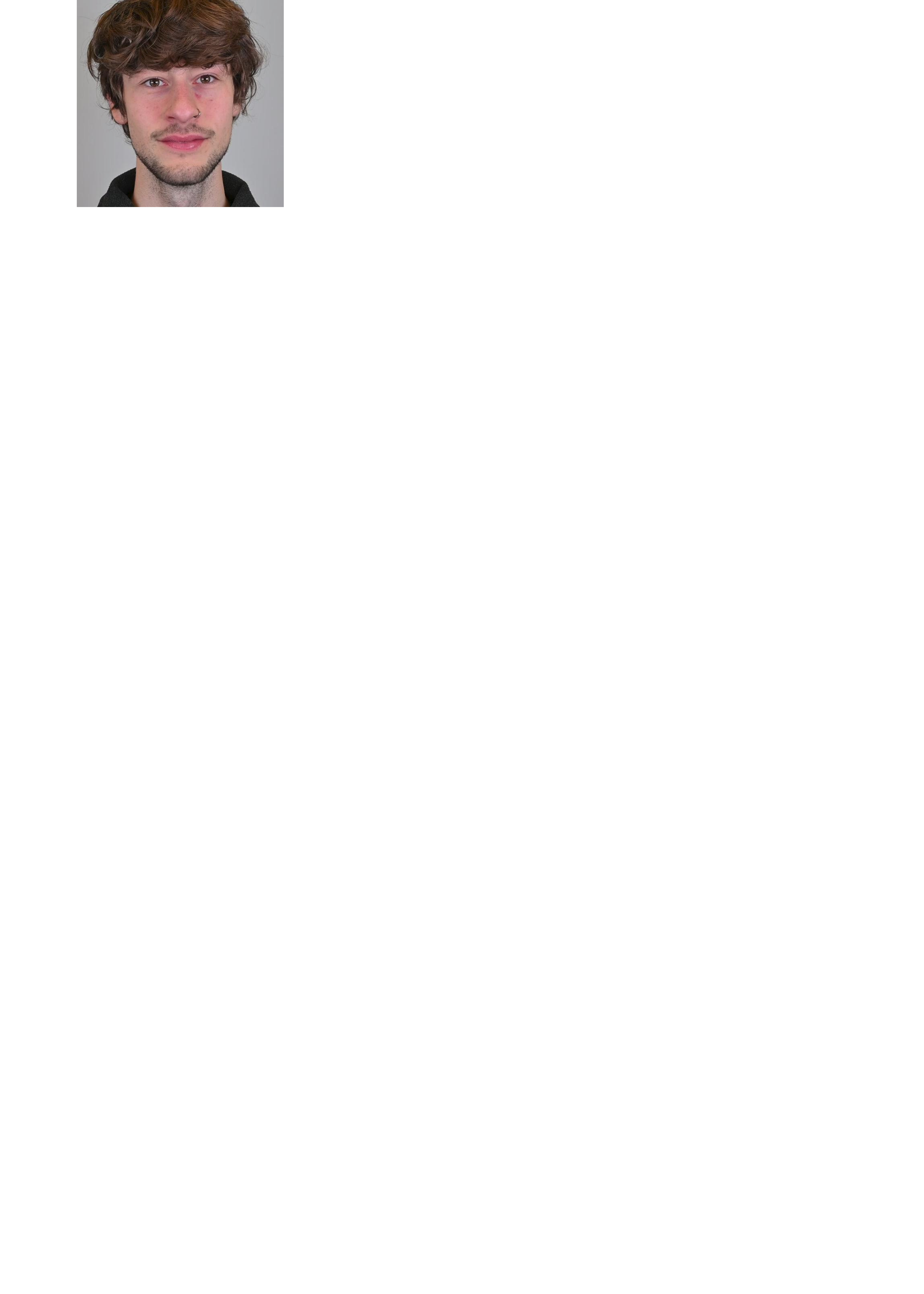
    \caption{Comparison of \textbf{3D reconstructions} from different distillation procedures. 
    *: Replacing the method's 2D prior with our model for fair comparison.}
    \label{fig:distillation_comparison}
\end{figure}

\begin{table}[t]
    \centering
    \resizebox{\linewidth}{!}{
    \begin{tabular}{l|cccc}
        & LPIPS $\downarrow$ & SSIM $\uparrow$ & PSNR $\uparrow$ & MSE $\downarrow$ \\
        \hline
        ProlificDreamer*~\cite{wang2023prolificdreamer} &                      0.43 &                      0.52 &                      15.6 &                      0.028 \\
ImageDream*~\cite{wang2023imagedream}     &  \cellcolor{tabthird}0.25 &                      0.79 &  \cellcolor{tabthird}20.2 &  \cellcolor{tabthird}0.011 \\
Ours, Stage 1 Only          &                      0.27 &  \cellcolor{tabfirst}0.83 &                      20.1 &                      0.012 \\
Ours, Stage 2 Only         &  \cellcolor{tabfirst}0.18 &  \cellcolor{tabthird}0.81 &  \cellcolor{tabfirst}22.0 &  \cellcolor{tabfirst}0.007 \\
\hline
Ours             & \cellcolor{tabsecond}0.19 & \cellcolor{tabsecond}0.82 & \cellcolor{tabsecond}21.5 & \cellcolor{tabsecond}0.008 \\

    \end{tabular}
    }
    \caption{Quantitative comparison of our \textbf{3D distillation} approach. *: Replacing the method's 2D prior with our model for fair comparison.}
    \label{tab:quant_comparison_distillation}
\end{table}

Note that the very high CSIM identity similarity score of X-Portrait for the cross-reenactment scenario is misleading. We found that for extreme pose changes, X-Portrait tends to reproduce the reference image without adopting the driving pose (see row 3 of \Cref{fig:2d_prior_comparison}). 
Since for the cross-reenactment scenario, we calculate CSIM between the reference image and the prediction, this artifact results in a significantly overestimated CSIM score. 
This effect is confirmed by the comparatively high pose reconstruction error (APD) of X-Portrait in \Cref{tab:quant_comparison}. 
Our method excels in high-fidelity synthesis and identity preservation while being robust w.r.t. extreme expressions and poses both in the reference and driving image.  
%


\subsection{Evaluation of 3D Distillation}

We compare our novel distillation approach against two state-of-the-art baselines: \emph{ImageDream}~\cite{wang2023imagedream} and \emph{ProlificDreamer}~\cite{wang2023prolificdreamer}.
\emph{ImageDream}~\cite{wang2023imagedream} uses a diffusion prior that predicts consistent multi-view images given a reference image and exploits this prior to perform multi-view score distillation sampling.   
\emph{ProlificDreamer}~\cite{wang2023prolificdreamer} generalizes score distillation sampling to variational score distillation by treating the NeRF renderings as random variables approximated by a pose-conditioned diffusion model that is fine-tuned on the NeRF renderings during distillation.
For ImageDream, we use the official code base, for ProlificDreamer we use the threestudio implementation \cite{threestudio2023}.
For a fair comparison, we replace the 2D diffusion priors of both baselines with our own prior.
We further compare against versions of our method that only use Stage 1 and Stage 2 respectively. 
Note that the Stage-2-only setting is similar to Cat3D~\cite{gao2024cat3d}. 
However, Cat3D only considers a novel view synthesis scenario where reference images of the same scene are given. 
Instead, in our scenario the reference and output images differ drastically due to strong pose and expression changes.  

\Cref{fig:distillation_comparison} presents a qualitative comparison of the distillation procedures on samples from the NerSemble validation set and our self-captured \textit{Joker benchmark} with in-the-wild scenarios.
We observe that ProlificDreamer exhibits instabilities during distillation. They are caused by inaccuracies in the fine-tuned diffusion prior that approximates the probability distribution of the NeRF renderings. 
ImageDream converges more stably, yet artifacts remain since even at the end of the distillation procedure, denoising timesteps are sampled randomly including high noise levels leading to inaccurate estimates of the target images for the optimization of the NeRF. 
These artifacts are resolved using the Stage~1 optimization of our approach that follows a deterministic denoising schedule.
However, using only Stage 1 yields blurry synthesis results. 
As discussed in \Cref{sec:distillation}, the low-frequency bias of Stage 1 stems from a repeated interplay between NeRF optimization and target image prediction.
When performing Stage 2 only, we suppress this effect and can distill a NeRF with high-frequency details. However, note that in Stage 2 all target images are generated at once and the NeRF is optimized against fixed targets. The inconsistencies in these images cause synthesis artifacts like semi-transparencies and misalignment artifacts in the eyes and around the silhouette.  
Please refer to the suppl. video for a dynamic comparison of the distillation results.
In \Cref{tab:quant_comparison_distillation}, we also perform a quantitative comparison on 30 samples from the NeRSemble validation set which provides multi-view ground truth images. 
Our distillation approach consistently improves over the baselines.
We observe that while using both stages of our method qualitatively yielded the best combination of view consistency and high-frequency detail, using only Stage 2 even improves the scores slightly, however, at the cost of view consistency. 
Please refer to the suppl. video for a dynamic visualization of this effect.
\Cref{sec:suppl_distillation_ablation} further provides an evaluation of the impact of the classifier-free guidance scale and the ratio between Stage 1 and Stage 2.
We observe that our distillation generates plausible geometry and generalizes well to challenging out-of-distribution samples (see \Cref{fig:ood}).

\subsection{Text-Guided Expression Synthesis}

\Cref{fig:text_control} demonstrates the effectiveness of using text prompts to control the 3D reconstruction and disambiguate the control through 3DMM parameters:
Each row presents two reconstruction results that use the same reference image and 3DMM parameters but different text prompts. 
We find that text prompts provide an intuitive control mechanism to specify the target emotion and tongue articulation. 
Please refer to \Cref{fig:2dprior_ablation} and \Cref{tab:ablation_2d_prior} for comparisons against a model without text control.

\subsection{Limitations}
\begin{figure}[t]
    \centering
    \def\svgwidth{\linewidth}
\begingroup%
  \makeatletter%
  \providecommand\color[2][]{%
    \errmessage{(Inkscape) Color is used for the text in Inkscape, but the package 'color.sty' is not loaded}%
    \renewcommand\color[2][]{}%
  }%
  \providecommand\transparent[1]{%
    \errmessage{(Inkscape) Transparency is used (non-zero) for the text in Inkscape, but the package 'transparent.sty' is not loaded}%
    \renewcommand\transparent[1]{}%
  }%
  \providecommand\rotatebox[2]{#2}%
  \newcommand*\fsize{\dimexpr\f@size pt\relax}%
  \newcommand*\lineheight[1]{\fontsize{\fsize}{#1\fsize}\selectfont}%
  \ifx\svgwidth\undefined%
    \setlength{\unitlength}{1955.99997693bp}%
    \ifx\svgscale\undefined%
      \relax%
    \else%
      \setlength{\unitlength}{\unitlength * \real{\svgscale}}%
    \fi%
  \else%
    \setlength{\unitlength}{\svgwidth}%
  \fi%
  \global\let\svgwidth\undefined%
  \global\let\svgscale\undefined%
  \makeatother%
  \scriptsize
  \begin{picture}(1,0.631994)%
    \lineheight{1}%
    \setlength\tabcolsep{0pt}%
    \put(0,0){\includegraphics[width=\unitlength,page=1]{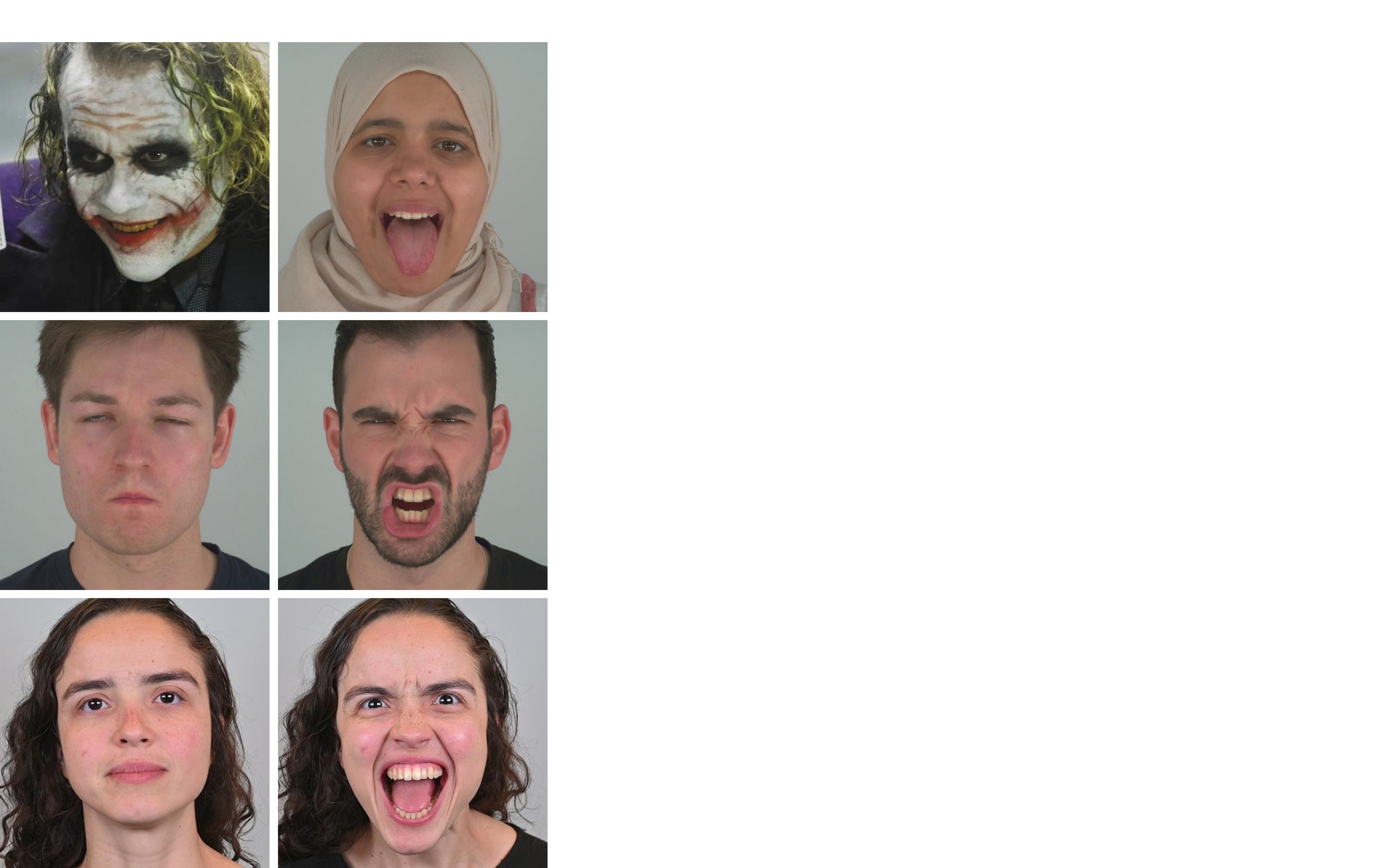}}%
    \put(0.09825137,0.61495511){\makebox(0,0)[t]{\lineheight{1.25}\smash{\begin{tabular}[t]{c}Reference Image\end{tabular}}}}%
    \put(0.30070536,0.61495511){\makebox(0,0)[t]{\lineheight{1.25}\smash{\begin{tabular}[t]{c}Driving Image\end{tabular}}}}%
    \put(0.70537169,0.61495511){\makebox(0,0)[t]{\lineheight{1.25}\smash{\begin{tabular}[t]{c}3D Reconstruction\end{tabular}}}}%
    \put(0,0){\includegraphics[width=\unitlength,page=2]{failure_cases.pdf}}%
  \end{picture}%
\endgroup%

    \caption{Failure cases of our method.}
    \label{fig:failure_cases}
\end{figure}

\Cref{fig:failure_cases} visualizes failure cases of our method. 
We observe that implausible colors may be synthesized for challenging out-of-distribution samples in face regions that are not visible in the reference image. 
In rare cases, we find that even for in-distribution samples the high cfg value (19.0) during distillation causes unnatural colorizations. Please refer to \Cref{sec:suppl_distillation_ablation} for an ablation study on this parameter during distillation. 
For samples with a uniform background, we observe a tendency of our model to project them to the NerSemble lighting setting: compare rows 2 (NeRSemble) and 3 (Joker benchmark) of \Cref{fig:failure_cases}. 
Lastly, particularly for dark curly hair, we observe a low-frequency bias. While using only Stage 2 of our distillation procedure can reduce this effect, this comes at the cost of reduced 3D consistency. 
Further, note that we only consider static scenes in our work. 
Extending it to 4D avatar synthesis is a fascinating topic for future research.

\section{Conclusion}

We introduced \emph{Joker}, a novel method for conditional 3D human head synthesis with extreme expressions from a single reference image.
Based on control through 3DMM parameters and text prompts, our method produces high-quality results and generalizes well to out-of-distribution samples.
The foundation of this approach is a 2D diffusion-based prior which is learned on in-the-wild imagery of human faces.
We leverage this prior to progressively distill a 3D volumetric representation of the target subject with a different facial expression.
The textual description allows us to specify the facial expression state beyond the parameters of the 3DMM, including subtle emotional changes, as well as extreme expressions with protruding tongue.
We believe that \textit{Joker} is a stepping stone for creating high-resolution 3D content of people with a high degree of identity preservation and emotional expressiveness.
\section{Acknowledgements}
This project has received funding from the Max Planck ETH Center for Learning Systems (CLS). Egor Zakharov was funded by the “AI-PERCEIVE” 2021 ERC Consolidator Grant. Further, we would like to thank Phong Tran and Balamurugan Thambiraja for their valuable feedback.  

{
    \small
    \bibliographystyle{ieeenat_fullname}
    \bibliography{main}
}

\appendix

\definecolor{tabfirst}{rgb}{1, 0.7, 0.7} 
\definecolor{tabsecond}{rgb}{1, 0.85, 0.7} 
\definecolor{tabthird}{rgb}{1, 1, 0.7} 

\twocolumn[{%
    \renewcommand\twocolumn[1][]{#1}%
    \maketitlesupplementary
    \begin{center}
        \centering
        \captionsetup{type=figure}
        \vspace{-0.5cm}
        \includegraphics[width=1.\textwidth]{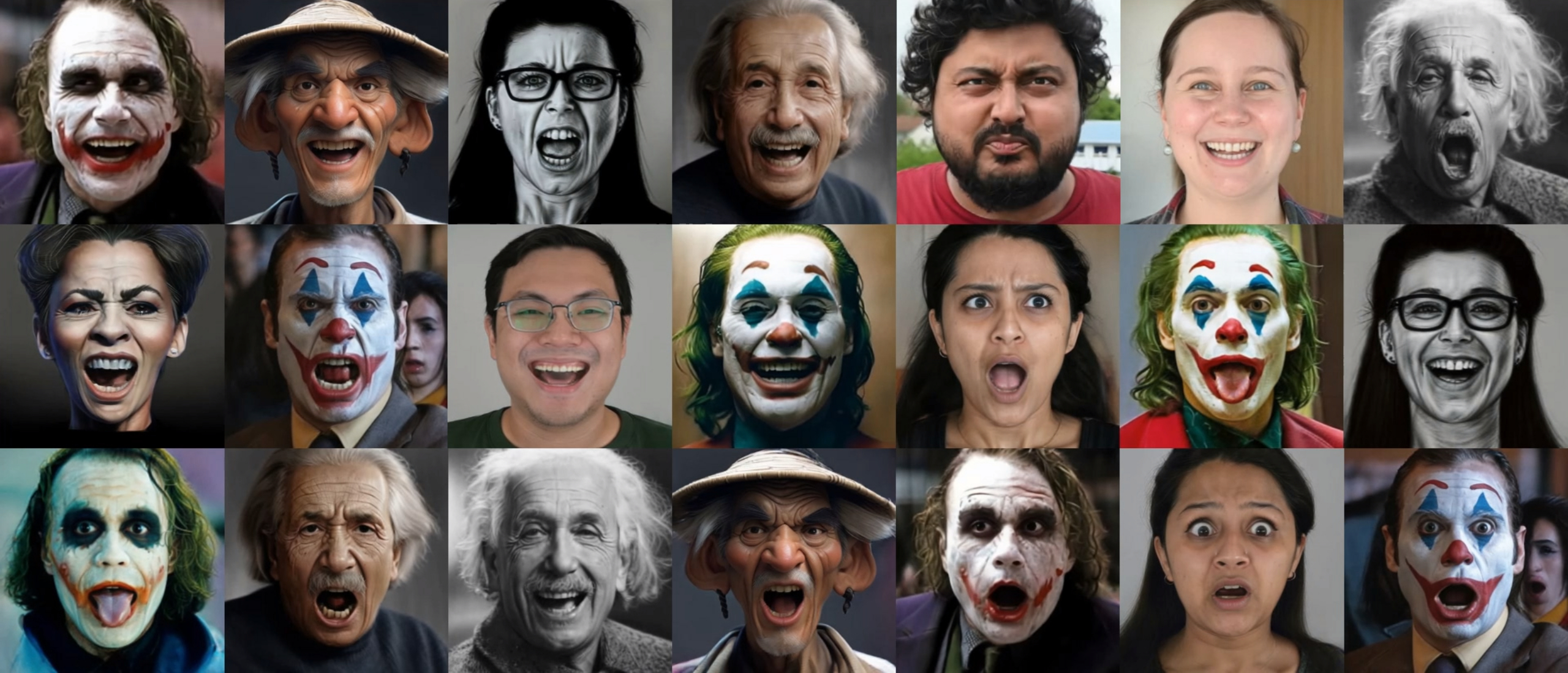}
        \caption{Further 3D reconstructions of our method on a diverse set of examples including out-of-distribution samples. 
        }
        \label{fig:teaser}
    \end{center}%
}]


\section{Datasets}
\label{sec:suppl_datasets}
As described in the main paper, we use the datasets CelebV-Text~\cite{yu2022celebvtext} and NeRSemble~\cite{kirschstein2023nersemble} and generate new annotations and metadata to train our model.
In addition, we recorded challenging samples for validating the synthesis of extreme facial expressions in an in-the-wild setting; this benchmark is referred to as \textit{Joker benchmark}.

\paragraph{CelebV-Text Dataset}
The CelebV-Text Dataset \cite{yu2022celebvtext} is a large-scale facial text-video dataset containing 70k facial video clips from the internet with a total length of 279 hours. 
We first filter out videos of low quality by discarding samples with a HyperIQA score~\cite{hyperiqa} of less than 40. 
Second, we filter for videos with extreme and diverse poses and expressions. For that, we use an off-the-shelf model\footnote{\url{https://github.com/radekd91/inferno/tree/master/inferno_apps/FaceReconstruction}}~\cite{danvevcek2022emoca} to annotate the video frames with 3DMM parameters and select the frames with the highest expressiveness and diversity.
The filtered images are cropped following the alignment procedure of \cite{deep3dfacerecon} and automatically annotated with BFM parameters and text captions using Deep3DFaceRecon~\cite{deep3dfacerecon} and Blip2~\cite{blip2}.
Samples for which the 3DMM parameters estimation fails and with implausible captions are discarded. 

We select 50k samples for training and 2.5k for evaluation. 
Reference images are randomly sampled from the same sequence as the target image, weighted by the relative distance in pose and expression. 
To avoid identity overlap between the training and validation sets, we use an off-the-shelf face recognition network \cite{deng2019arcface} and enforce an identity similarity score of less than $0.4$ between each validation sample and its closest training sample.
The automatically generated annotations and metadata will be made publicly available to the research community.

\paragraph{NeRSemble Dataset}
The NeRSemble dataset \cite{kirschstein2023nersemble} is a multi-view portrait video dataset containing 4734 recordings of 222 subjects captured with 16 machine vision cameras. The subjects perform a wide set of extreme expressions in an environment with uniform lighting and background.
We follow the same procedure as for CelebV-Text for sample filtering, image cropping, and annotation. 
Further, we assign a higher sampling ratio to the samples for which the automatically generated caption contains the keyword "tongue" because such samples are sparse in the CelebV-Text dataset. 
Note that we only create the image captions for the frontal images and reuse them for the other multi-view images. 
Reference images are randomly sampled from images showing the same subject as the target image but with a different expression and captured from a different camera.  
We split the dataset into 199 subjects for training and 23 for validation and automatically selected 2,000 and 2,500 frames, respectively. 

\begin{figure*}[t]
    \centering
      \def\svgwidth{\linewidth}
\begingroup%
  \makeatletter%
  \scriptsize
  \providecommand\color[2][]{%
    \errmessage{(Inkscape) Color is used for the text in Inkscape, but the package 'color.sty' is not loaded}%
    \renewcommand\color[2][]{}%
  }%
  \providecommand\transparent[1]{%
    \errmessage{(Inkscape) Transparency is used (non-zero) for the text in Inkscape, but the package 'transparent.sty' is not loaded}%
    \renewcommand\transparent[1]{}%
  }%
  \providecommand\rotatebox[2]{#2}%
  \newcommand*\fsize{\dimexpr\f@size pt\relax}%
  \newcommand*\lineheight[1]{\fontsize{\fsize}{#1\fsize}\selectfont}%
  \ifx\svgwidth\undefined%
    \setlength{\unitlength}{3192.00029989bp}%
    \ifx\svgscale\undefined%
      \relax%
    \else%
      \setlength{\unitlength}{\unitlength * \real{\svgscale}}%
    \fi%
  \else%
    \setlength{\unitlength}{\svgwidth}%
  \fi%
  \global\let\svgwidth\undefined%
  \global\let\svgscale\undefined%
  \makeatother%
  \begin{picture}(1,0.387218)%
    \lineheight{1}%
    \setlength\tabcolsep{0pt}%
    \put(0,0){\includegraphics[width=\unitlength,page=1]{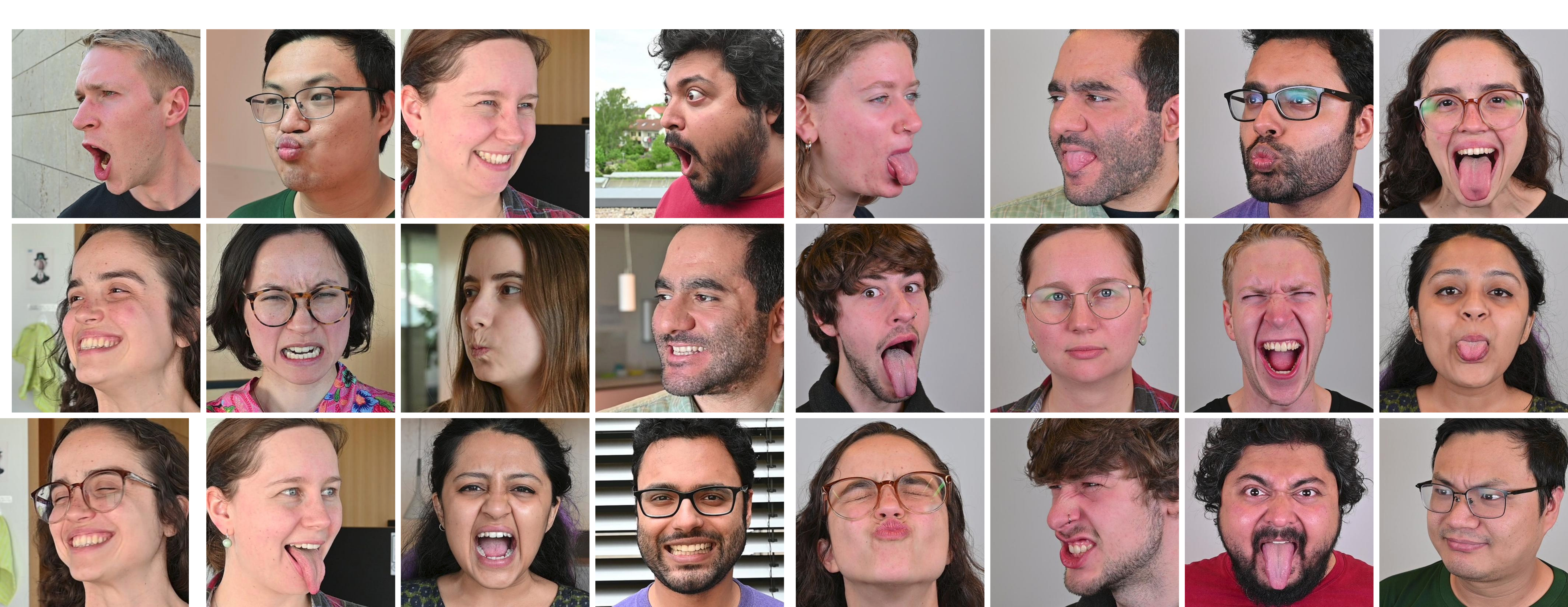}}%
    \put(0.25409396,0.37788467){\makebox(0,0)[t]{\lineheight{1.25}\smash{\begin{tabular}[t]{c}In-the-Wild Capture\end{tabular}}}}%
    \put(0.75359401,0.37788467){\makebox(0,0)[t]{\lineheight{1.25}\smash{\begin{tabular}[t]{c}Studio Capture\end{tabular}}}}%
    \put(0,0){\includegraphics[width=\unitlength,page=2]{joker_benchmark_samples.pdf}}%
  \end{picture}%
\endgroup%

    \caption{Random samples from our \emph{Joker benchmark}. The samples contain in-the-wild scenes with natural backgrounds and lighting and studio scenes with uniform backgrounds and lighting. }
    \label{fig:joker_benchmark}
\end{figure*}

\paragraph{Joker Benchmark}
Evaluating our method on the validation sets of CelebV-Text and NeRSemble alone is insufficient: CelebV-Text only contains moderately extreme expressions, and NeRSemble is restricted to uniform lighting and background scenarios. 
For this reason, we captured the \emph{Joker benchmark} for the evaluation of extreme expression synthesis, which will be made publicly available to the research community. 
It provides monocular videos of 13 subjects performing extreme expressions both in in-the-wild scenarios, as well as in a lab environment with uniform lighting and background, see \Cref{fig:joker_benchmark}.
The subjects are of diverse ethnicity and equal gender parity (6 male, 7 female).
We apply the same alignment and annotation pipeline to the dataset as for CelebV-Text.

\begin{figure*}[t]
    \centering
    \def\svgwidth{\linewidth}
    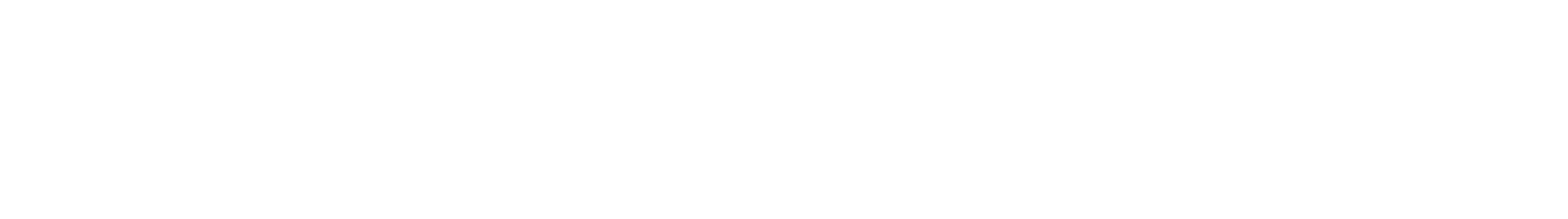
    \caption{Qualitative ablation study of our \textbf{2D prior}. Too small classifier-free guidance scale values (cfg) reduce the faithfulness of extreme expressions, while too high values cause oversaturation and artifacts. We find that cfg=3 yields the best trade-off. Not training the upsampling layers of the denoising UNet (\emph{"Frozen Denoising UNet"}) worsens identity preservation and synthesis quality in general. Dropping the text control disables tongue control since it is not represented in the 3DMM. Similar effects occur when not fine-tuning on NeRSemble, since samples with visible tongue are underrepresented in CelebV-Text.}
    \label{fig:2dprior_ablation}
\end{figure*}

\begin{table*}[t]
    \centering
    \resizebox{\linewidth}{!}{
    \begin{tabular}{l|cccccccc|cccc}
         & \multicolumn{8}{c|}{Self-reenactment} & \multicolumn{4}{c}{Cross-reenactment} \\
         & \multicolumn{1}{c}{PSNR $\uparrow$} & \multicolumn{1}{c}{LPIPS $\downarrow$} & \multicolumn{1}{c}{SSIM $\uparrow$} & \multicolumn{1}{c}{FID $\downarrow$} & \multicolumn{1}{c}{CSIM $\uparrow$} & \multicolumn{1}{c}{AKD $\downarrow$} & \multicolumn{1}{c}{AED $\downarrow$} & \multicolumn{1}{c|}{APD $\downarrow$} & \multicolumn{1}{c}{FID $\downarrow$} & \multicolumn{1}{c}{CSIM $\uparrow$} & \multicolumn{1}{c}{AED $\downarrow$} & \multicolumn{1}{c}{APD $\downarrow$} \\
        \hline

Frozen Denoising UNet     &  \cellcolor{tabthird}18.11 &  \cellcolor{tabthird}0.227 &  \cellcolor{tabthird}0.603 &  \cellcolor{tabthird}8.38 &  \cellcolor{tabthird}0.58 &  \cellcolor{tabthird}0.0072 &  \cellcolor{tabthird}0.119 &  \cellcolor{tabthird}0.325 &  \cellcolor{tabthird}9.59 &                      0.54 &  \cellcolor{tabthird}0.223 &  \cellcolor{tabthird}0.494 \\
No Text Control &                      17.02 &                      0.259 &                      0.566 &                      13.46 &                      0.56 &                      0.0132 &                      0.148 &                      0.380 &                      14.68 &  \cellcolor{tabthird}0.55 &                      0.249 &                      0.530 \\
No Finetuning on NerSemble         &  \cellcolor{tabfirst}18.72 &  \cellcolor{tabfirst}0.210 &  \cellcolor{tabfirst}0.622 & \cellcolor{tabsecond}8.15 & \cellcolor{tabsecond}0.61 &  \cellcolor{tabfirst}0.0061 &  \cellcolor{tabfirst}0.109 & \cellcolor{tabsecond}0.310 & \cellcolor{tabsecond}9.00 &  \cellcolor{tabfirst}0.58 & \cellcolor{tabsecond}0.221 &  \cellcolor{tabfirst}0.486 \\

\hline
\textbf{Ours}               & \cellcolor{tabsecond}18.63 & \cellcolor{tabsecond}0.212 & \cellcolor{tabsecond}0.619 &  \cellcolor{tabfirst}7.57 &  \cellcolor{tabfirst}0.62 & \cellcolor{tabsecond}0.0067 & \cellcolor{tabsecond}0.110 &  \cellcolor{tabfirst}0.306 &  \cellcolor{tabfirst}8.48 & \cellcolor{tabsecond}0.57 &  \cellcolor{tabfirst}0.220 & \cellcolor{tabsecond}0.489 \\

    \end{tabular}
    }
    \caption{Quantitative ablation study of the design choices of our \textbf{2D prior}.}
    \label{tab:ablation_2d_prior}
\end{table*}

\begin{table*}[th!]
    \centering
    \resizebox{\linewidth}{!}{
    \begin{tabular}{l|cccccccc|cccc}
         & \multicolumn{8}{c|}{Self-reenactment} & \multicolumn{4}{c}{Cross-reenactment} \\
         & \multicolumn{1}{c}{PSNR $\uparrow$} & \multicolumn{1}{c}{LPIPS $\downarrow$} & \multicolumn{1}{c}{SSIM $\uparrow$} & \multicolumn{1}{c}{FID $\downarrow$} & \multicolumn{1}{c}{CSIM $\uparrow$} & \multicolumn{1}{c}{AKD $\downarrow$} & \multicolumn{1}{c}{AED $\downarrow$} & \multicolumn{1}{c|}{APD $\downarrow$} & \multicolumn{1}{c}{FID $\downarrow$} & \multicolumn{1}{c}{CSIM $\uparrow$} & \multicolumn{1}{c}{AED $\downarrow$} & \multicolumn{1}{c}{APD $\downarrow$} \\
        \hline

Ours, cfg=1.0  & \cellcolor{tabsecond}18.49 & \cellcolor{tabsecond}0.216 &  \cellcolor{tabthird}0.611 &  \cellcolor{tabthird}9.95 &  \cellcolor{tabfirst}0.618 &  \cellcolor{tabfirst}0.00667 &  \cellcolor{tabfirst}0.109 & \cellcolor{tabsecond}0.310 &  \cellcolor{tabthird}11.03 &  \cellcolor{tabfirst}0.567 & \cellcolor{tabsecond}0.2203 & \cellcolor{tabsecond}0.490 \\
\hline
$\textbf{Ours, cfg=3.0}$  &  \cellcolor{tabfirst}18.63 &  \cellcolor{tabfirst}0.212 &  \cellcolor{tabfirst}0.619 &  \cellcolor{tabfirst}7.57 & \cellcolor{tabsecond}0.616 & \cellcolor{tabsecond}0.00669 & \cellcolor{tabsecond}0.110 &  \cellcolor{tabfirst}0.306 &  \cellcolor{tabfirst}8.48 & \cellcolor{tabsecond}0.566 &  \cellcolor{tabfirst}0.2201 &  \cellcolor{tabfirst}0.489 \\
\hline
Ours, cfg=6.0  &  \cellcolor{tabthird}18.40 &  \cellcolor{tabthird}0.221 & \cellcolor{tabsecond}0.618 & \cellcolor{tabsecond}8.12 &  \cellcolor{tabthird}0.594 &  \cellcolor{tabthird}0.00690 &  \cellcolor{tabthird}0.116 &  \cellcolor{tabthird}0.318 & \cellcolor{tabsecond}9.05 &  \cellcolor{tabthird}0.548 &  \cellcolor{tabthird}0.2224 &  \cellcolor{tabthird}0.491 \\
Ours, cfg=10.0 &                      18.10 &                      0.234 &  \cellcolor{tabthird}0.611 &                      10.58 &                      0.572 &                      0.00712 &                      0.122 &                      0.329 &                      11.70 &                      0.528 &                      0.2245 &                      0.493

    \end{tabular}
    }
    \caption{Quantitative ablation study of the impact of classifier-free guidance scale (cfg) on our \textbf{2D prior}.}
    \label{tab:ablation_2dprior_cfg}
\end{table*}


\section{Description of the baseline methods}
\label{sec:suppl_baselines}

\emph{VOODOO 3D}~\cite{tran2023voodoo} finetunes a pretrained model \cite{trevithick2023real} to lift the reference image into 3D and trains a model to transfer expressions between the 3D representations of the driving and the reference subject. 
\emph{VOODOO XP}~\cite{tran2024voodooxp} similarly to VOODOO 3D also leverages 3D lifting but learns an expression encoder in an end-to-end fashion to provide fine-grained expression control. 
\emph{Real3D-Portrait}~\cite{ye2024real3dportrait}  combines an image-to-plane model with a tri-plane motion adapter to synthesize 3D talking head avatars that can be controlled via audio or 3DMM parameters. 
\emph{Portrait4D-v2}~\cite{deng2024portrait4dv2} combines a modified EG3D~\cite{eg3d} pipeline with a control mechanism through the FLAME 3DMM \cite{FLAME:SiggraphAsia2017}. 
\emph{GOHA}~\cite{goha} uses a 3DMM to control facial expressions by mapping 3DMM parameters to residuals of a tri-plane representation of the face. 
\emph{AniFaceDiff}~\cite{anifacediff} follows a similar approach as our method, yet instead of using a ControlNet, they encode normal maps of FLAME~\cite{FLAME:SiggraphAsia2017} through stacked 2D convolutions and directly add them to the noisy input latents. Further, they don't use text control but apply cross-attention to features extracted from the FLAME parameters. 
\emph{X-Portrait}~\cite{xie2024xportrait}  also follows a similar approach as our method. In contrast to our method, however, they don't utilize text and 3DMM as inputs. Instead, they use patches of the driving image as input to the ControlNet. To avoid identity leakage during training, X-Portrait uses a pre-trained facial reenactment method to generate them. 

Note that for the baselines Real3DPortrait, AniFaceDiff, and X-Portrait, we use the renderings of our method to obtain the dynamic camera sweep results presented in the suppl. video. 
For the other baselines, we directly use the ground truth camera parameters for rendering. 

\section{Additional Experiments}
\label{sec:suppl_additional_experiments}

\subsection{Ablation Study of Our 2D Prior}
We ablate the design choices of our 2D prior in \Cref{tab:ablation_2d_prior} and \Cref{fig:2dprior_ablation}. 
In contrast to X-Portrait, we unfreeze the upsampling blocks of our denoising UNet and find that this consistently improves all metrics. Qualitatively, we observe particularly significant improvements for identity preservation under extreme expression changes~(see results 'Frozen Denoising UNet' in \Cref{fig:2dprior_ablation}). 
Removing the text control from our method during training and inference significantly worsens all metrics~(see results for 'No Text Control'). Qualitatively, we observe that extreme expressions, most prominently tongue articulations, cannot be controlled through 3DMM parameters alone, which explains the observed deterioration of the evaluation scores.
Note that none of the existing methods can leverage text for avatar control.
We found that fine-tuning our model on a mixture of NerSemble and CelebV-Text after pretraining on CelebV-Text greatly helps in synthesizing tongue articulations (see last column of \Cref{fig:2dprior_ablation}) since these samples are underrepresented in CelebV-Text.
However, the quantitative scores slightly deteriorate. We attribute this to a slight overfitting effect on the lighting situation of NeRSemble which causes predictions on samples with uniform backgrounds to have a bias toward this particular lighting setting. 
We also evaluate the impact of the classifier-free guidance scale (cfg) on our 2D prior in \Cref{fig:2dprior_ablation} and \Cref{tab:ablation_2dprior_cfg}. We found that too small values reduce the faithfulness of extreme expressions while too high values cause oversaturation artifacts. We found that cfg=3 is a good compromise and also achieves the best FID, PSNR, LPIPS, and SSIM scores in the quantitative self-reenactment evaluation.

\subsection{Collapse of Dynamic-Target 
Distillation Approaches for Small Noise Levels}

\begin{figure}[t]
    \centering
    \def\svgwidth{\linewidth}
    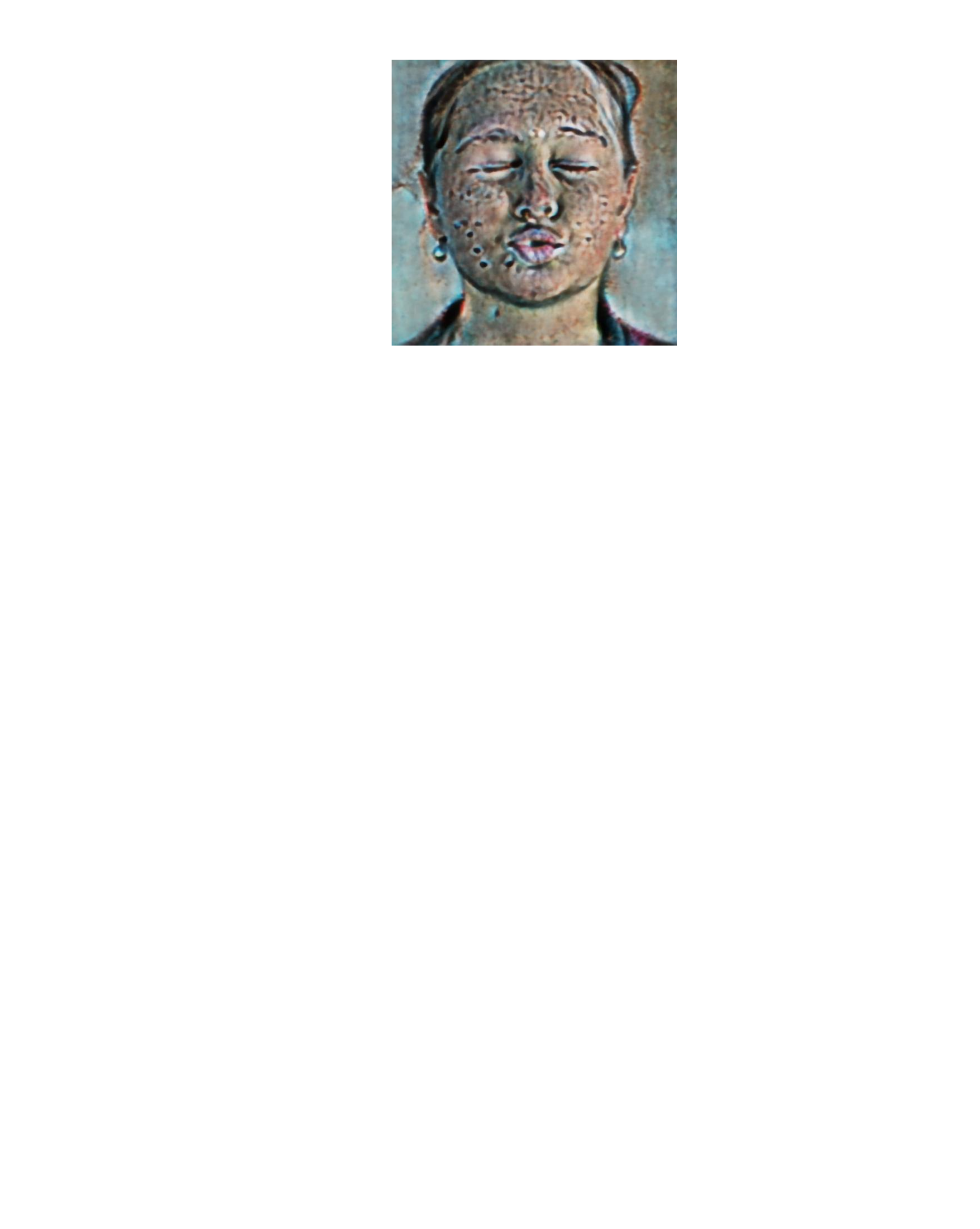
    \caption{Divergent behavior of score-distillation sampling (SDS) for small noise levels. Typically, SDS-based methods like ImageDream~\cite{wang2023imagedream} ensure that even towards the end of the distillation procedure high noise levels are sampled, i.e. $t\sim[0.02, 0.5]$ (see column 2). This bottlenecks the fidelity of the 3D reconstructions and causes artifacts for high cfg values. We found that SDS diverges when not ensuring high-noise levels towards the end of distillation, i.e. $t\sim [0.02, 0.02]$ (\emph{"without capped tmax"}, column~1). Our 2-staged approach with deterministic noise levels is able to overcome this limitation (3rd column).}
    \label{fig:sds_divergence}
\end{figure}

In the main paper, we found that our distillation approach yields better reconstruction results than methods like ImageDream~\cite{wang2023imagedream}, which update the target images at each NeRF optimization step (=\emph{"dynamic target"}). 
We argue that this is because such approaches sample the noise levels randomly from a specified range, even at the last step of distillation. 
However, the predictions of the 2D prior at high noise levels typically lack details and exhibit artifacts, particularly for high cfg values. 
Their contribution to the optimization objective bottlenecks the quality of the distillation result. 
The natural question is if the negative impact of high noise level sampling can be avoided by annealing the upper bound of the sampled noise levels to zero (note that ImageDream caps it to at least 0.5 by default).
The result of this experiment is demonstrated in \Cref{fig:sds_divergence}.
We found that annealing the upper bound of the noise levels to zero makes the distillation diverge. 
The reason for this is that when performing score distillation sampling (SDS) on small noise levels only, supervision for the low-frequency features like the general shape and outline of the distilled scene is lacking because, at these low-noise levels, only high-frequency details are added by the diffusion prior while the rest is copied over from the input images.
However, minor inaccuracies in this process cause the coarse geometry of the 3D reconstruction to drift during the repeated SDS updates while the diffusion prior does not provide correcting gradient directions.
As a result, the 3D reconstruction diverges.
Only by also sampling high noise levels even at the end of the distillation procedure, guidance on the coarse scene geometry can be achieved, while coming at the cost of reconstruction fidelity.

\subsection{Ablations of Our 3D Distillation Procedure}
\label{sec:suppl_distillation_ablation}
\begin{figure*}[t]
    \centering
    \def\svgwidth{\linewidth}
    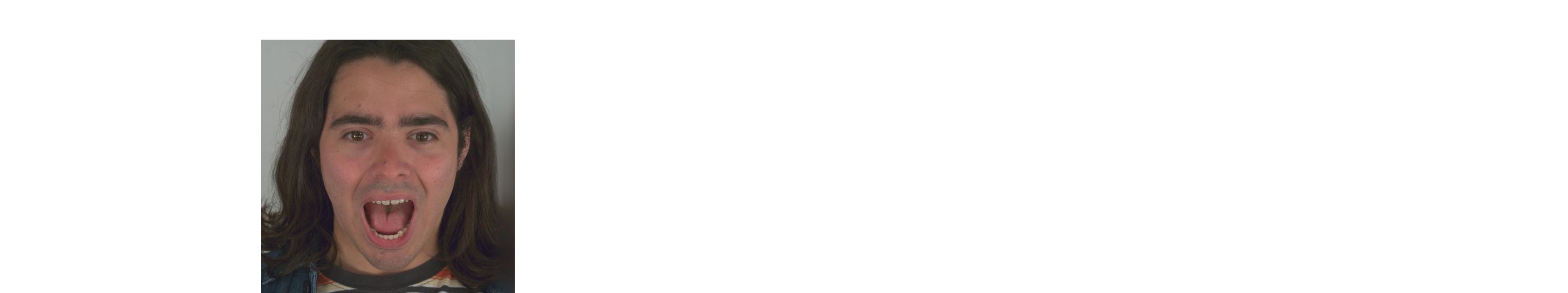
    \caption{Qualitative ablation study of classifier-free guidance scale (cfg) for our \textbf{3D distillation} procedure. Too small values produce blurry results, while too high values cause unnatural oversaturation. We found cfg=19.0 to be a good compromise and set it as the default for our method.}
    \label{fig:ablation_cfg_distillation}
\end{figure*}

\paragraph{Classifier-free guidance scale (cfg)}
\Cref{fig:ablation_cfg_distillation} and \Cref{tab:ablation_dist_cfg} ablate the impact of the cfg value during distillation. For the quantitative evaluation in \Cref{tab:ablation_dist_cfg}, we follow the same procedure as in the main paper. We find that too small cfg values ($\sim5$) produce blurry results while too high values ($\sim 30$) result in oversaturation. 
We chose cfg=19.0 and found that it yields plausible results of high quality without oversaturation effects. 

\paragraph{Ratio between Stage 1 \& 2}
\Cref{tab:ablation_dist_stageratios} provides a quantitative ablation study of the impact of the ratios between Stage 1 and Stage 2 during distillation. Please refer to the main paper for a qualitative comparison. 
We find that increasing the ratio of Stage 2 optimization improves high-frequency detail, the LPIPS score improves, yet comes at the cost of reduced consistency and semi-transparent artifacts, the structural similarity index measure (SSIM) worsens. 
We chose the ratio Stage 1 / Stage 2 of $60\% / 40\%$ as our default which we found to be a good trade-off between high-frequency details and consistency.

\begin{table}[t]
    \centering
    \resizebox{\linewidth}{!}{
    \begin{tabular}{l|cccc}
         & LPIPS $\downarrow$ & SSIM $\uparrow$ & PSNR $\uparrow$ & MSE $\downarrow$ \\
        \hline

        Ours, cfg=5.0  &  \cellcolor{tabthird}0.199 &  \cellcolor{tabfirst}0.84 &  \cellcolor{tabfirst}22.00 &  \cellcolor{tabfirst}0.0073 \\
Ours, cfg=10.0 & \cellcolor{tabsecond}0.193 & \cellcolor{tabsecond}0.83 & \cellcolor{tabsecond}21.77 & \cellcolor{tabsecond}0.0076 \\
\hline
\textbf{Ours, cfg=19.0} &  \cellcolor{tabfirst}0.191 &  \cellcolor{tabthird}0.82 &                      21.53 &                      0.0080 \\
\hline
Ours, cfg=30.0 &  \cellcolor{tabfirst}0.191 &                      0.81 &  \cellcolor{tabthird}21.60 &  \cellcolor{tabthird}0.0079

    \end{tabular}
    }
    \caption{Quantitative ablation study of the impact of classifier-free guidance scale (cfg) on our \textbf{3D distillation} procedure.}
    \label{tab:ablation_dist_cfg}
\end{table}

\begin{table}[t]
    \centering
    \resizebox{\linewidth}{!}{
    \begin{tabular}{c|cccc}
 Stage 1 / Stage 2    & LPIPS $\downarrow$ & SSIM $\uparrow$ & PSNR $\uparrow$ & MSE $\downarrow$ \\
\hline
        
100\%  / 0\%                  &                      0.27 &  \cellcolor{tabfirst}0.83 &                      20.1 &                      0.012 \\
80\% / 20\%                  &  \cellcolor{tabthird}0.21 & \cellcolor{tabsecond}0.82 &  \cellcolor{tabthird}20.6 &  \cellcolor{tabthird}0.010 \\
\hline
	$\textbf{60\% / 40\%}$ & \cellcolor{tabsecond}0.19 & \cellcolor{tabsecond}0.82 & \cellcolor{tabsecond}21.5 & \cellcolor{tabsecond}0.008 \\
\hline
30\% / 70\%                &  \cellcolor{tabfirst}0.18 &  \cellcolor{tabthird}0.81 &  \cellcolor{tabfirst}22.0 &  \cellcolor{tabfirst}0.007 \\
0\% / 100\%                  &  \cellcolor{tabfirst}0.18 &  \cellcolor{tabthird}0.81 &  \cellcolor{tabfirst}22.0 &  \cellcolor{tabfirst}0.007
        
    \end{tabular}}
    \caption{Quantitative ablation study of the ratios of Stage 1 and Stage 2 during our \textbf{3D distillation}. We use the ratio $60\%/40\%$ as the default for our method. While higher ratios of Stage 2 yield better LPIPS, we qualitatively found that it comes at the cost of less consistent reconstructions with semi-transparent artifacts (see main paper).}
    \label{tab:ablation_dist_stageratios}
\end{table}

\subsection{Qualitative Geometry Evaluation}
\begin{figure}[h]
    \centering
      \def\svgwidth{\linewidth}
\begingroup%
  \makeatletter%
  \scriptsize
  \providecommand\color[2][]{%
    \errmessage{(Inkscape) Color is used for the text in Inkscape, but the package 'color.sty' is not loaded}%
    \renewcommand\color[2][]{}%
  }%
  \providecommand\transparent[1]{%
    \errmessage{(Inkscape) Transparency is used (non-zero) for the text in Inkscape, but the package 'transparent.sty' is not loaded}%
    \renewcommand\transparent[1]{}%
  }%
  \providecommand\rotatebox[2]{#2}%
  \newcommand*\fsize{\dimexpr\f@size pt\relax}%
  \newcommand*\lineheight[1]{\fontsize{\fsize}{#1\fsize}\selectfont}%
  \ifx\svgwidth\undefined%
    \setlength{\unitlength}{1653.32802119bp}%
    \ifx\svgscale\undefined%
      \relax%
    \else%
      \setlength{\unitlength}{\unitlength * \real{\svgscale}}%
    \fi%
  \else%
    \setlength{\unitlength}{\svgwidth}%
  \fi%
  \global\let\svgwidth\undefined%
  \global\let\svgscale\undefined%
  \makeatother%
  \begin{picture}(1,0.47177569)%
    \lineheight{1}%
    \setlength\tabcolsep{0pt}%
    \put(0,0){\includegraphics[width=\unitlength,page=1]{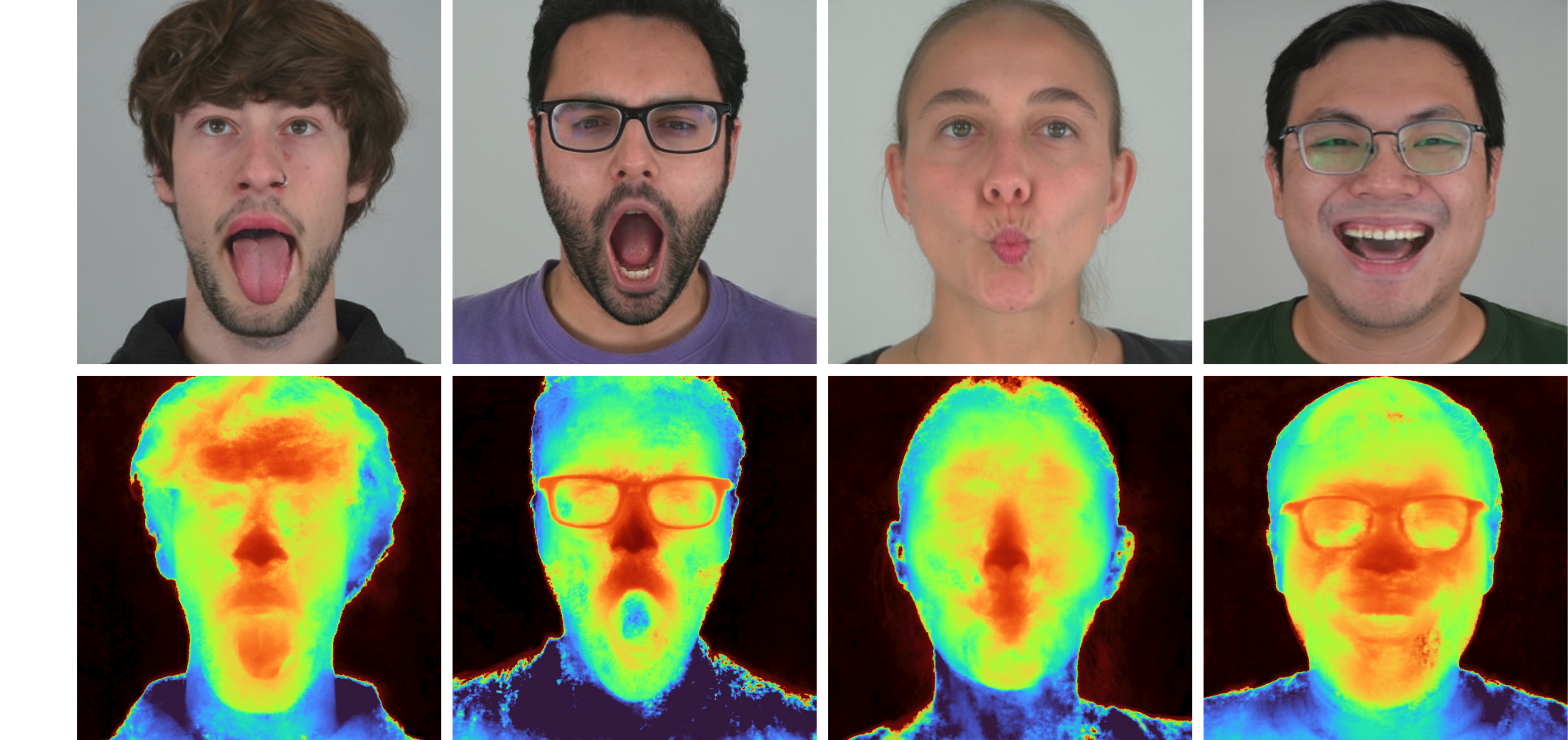}}%
    \put(0.02209555,0.35552817){\rotatebox{90}{\makebox(0,0)[t]{\lineheight{1.25}\smash{\begin{tabular}[t]{c}3D Reconstruction\end{tabular}}}}}%
    \put(0.02015817,0.1305416){\rotatebox{90}{\makebox(0,0)[t]{\lineheight{1.25}\smash{\begin{tabular}[t]{c}Depth\end{tabular}}}}}%
  \end{picture}%
\endgroup%

    \caption{Reconstructed Geometry.}
    \label{fig:depth}
\end{figure}

\Cref{fig:depth} qualitatively visualizes the depth maps of our 3D reconstructions. We observe that our distillation procedure yields plausible geometries with a distinct spatial separation of regions like nose, tongue, mouth cavities, and glasses.

\subsection{More Qualitative Comparisons of Our 2D Prior}
\begin{figure*}[t]
    \centering
      \def\svgwidth{\linewidth}
  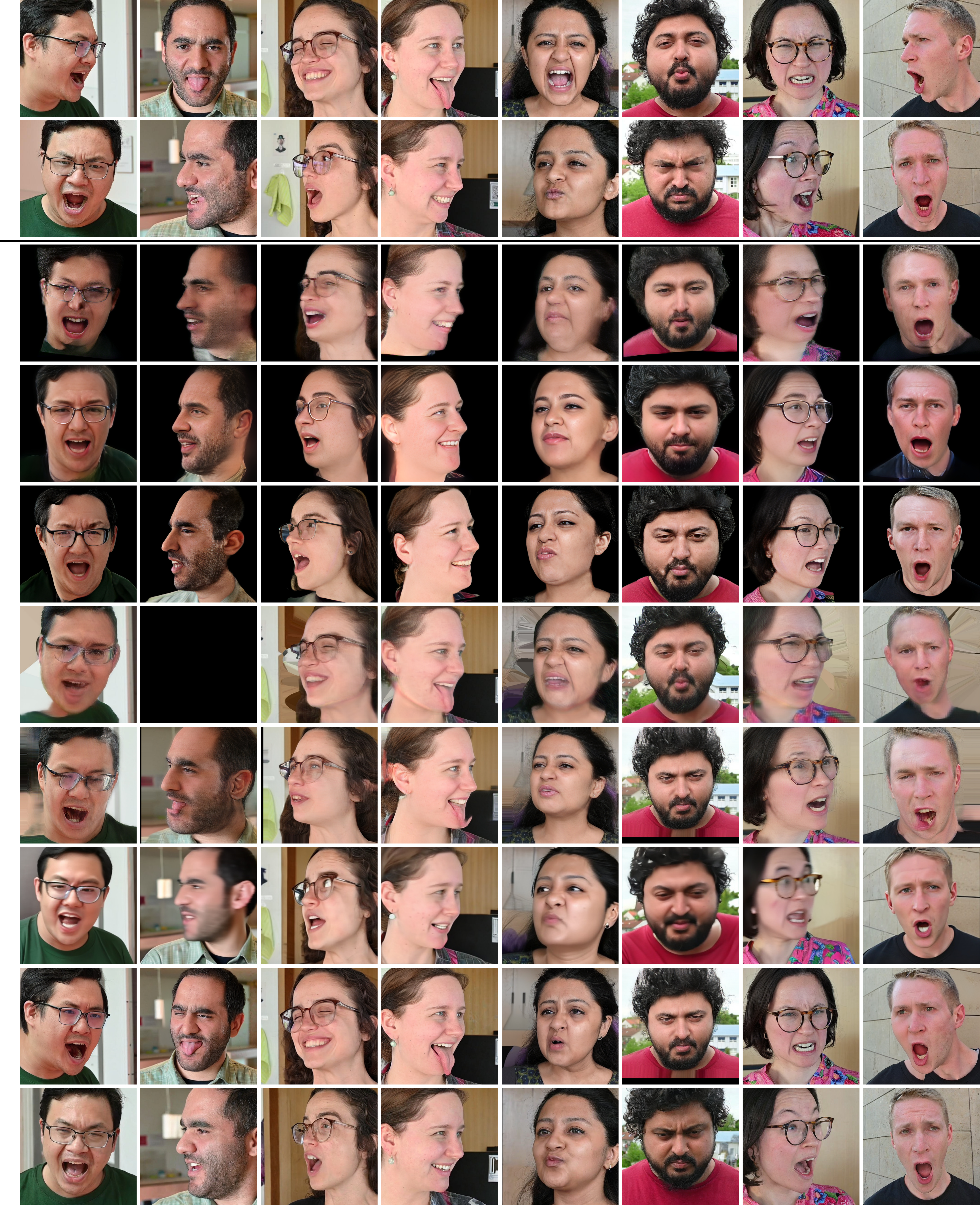
    \caption{Further qualitative comparisons of our \textbf{2D prior} in the self-reenactment scenario. For one sample, Real3D-Portrait's pose estimator failed, it is marked as a black tile. }
    \vspace{-1cm}
    \label{fig:2d_prior_comparison_large_self}
\end{figure*}

\begin{figure*}[t]
    \centering
      \def\svgwidth{\linewidth}
  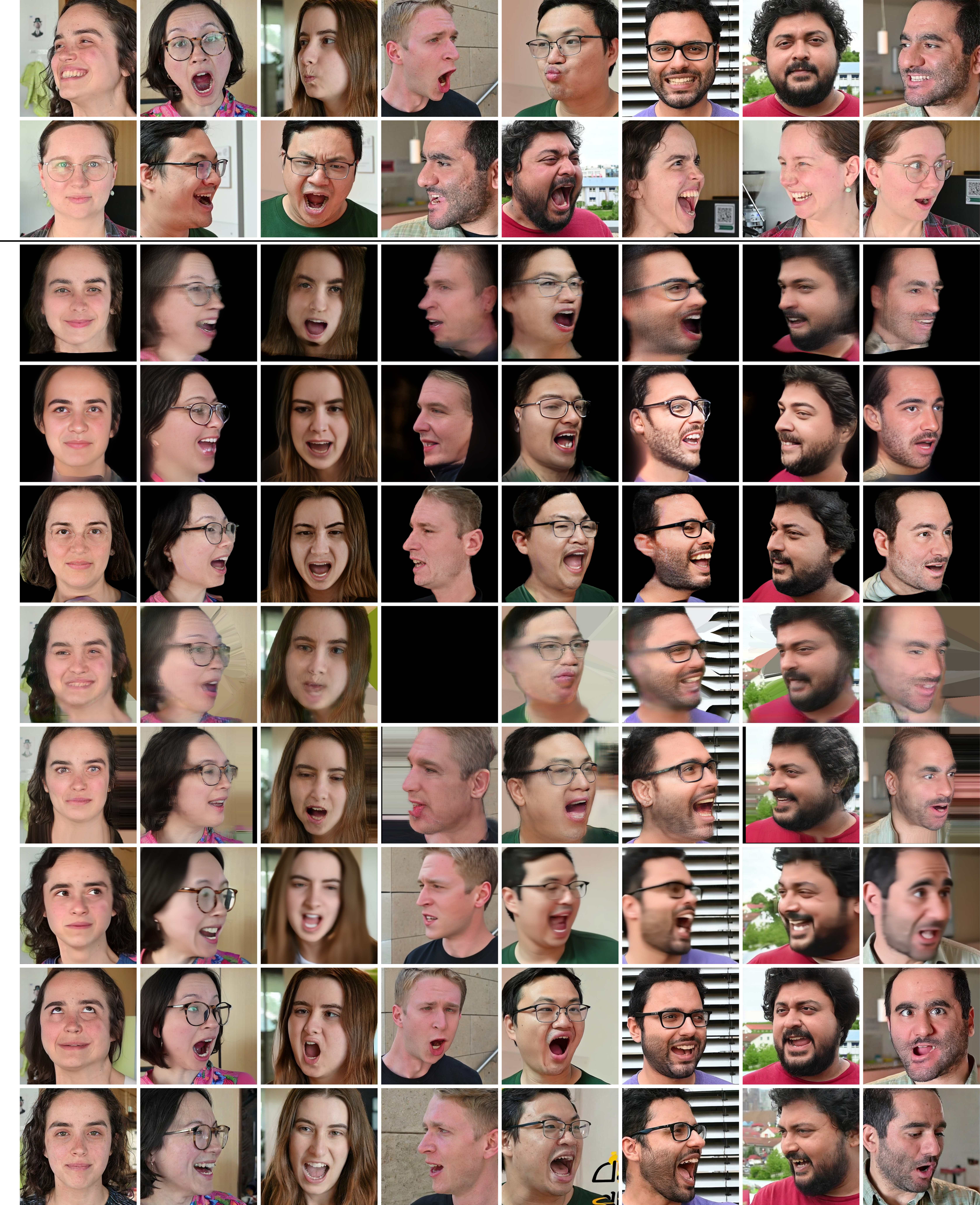
    \caption{Further qualitative comparisons of our \textbf{2D prior} in the cross-reenactment scenario. For one sample, Real3D-Portrait's pose estimator failed, it is marked as a black tile. }
    \vspace{-1cm}
    \label{fig:2d_prior_comparison_large_cross}
\end{figure*}

We provide additional qualitative comparisons of our 2D prior with all considered baselines in \Cref{fig:2d_prior_comparison_large_self} for self-reenactment and in \Cref{fig:2d_prior_comparison_large_cross} for cross-reenactment. 
As observed in the main paper, our 2D prior consistently outperforms all baselines.
It is remarkably robust w.r.t. extreme expressions and poses in the reference and the driving images and produces results with high identity alignment and synthesis quality even on very challenging samples. 

\section{Ethical Considerations}
Our method creates a photo-realistic 3D head reconstruction from a single reference image while providing control over the target pose and expression.
It is intended to advance 3D content generation for applications in telecommunications, movie production, and entertainment. 
Nevertheless, similar to previous work~\cite{xie2024xportrait, tran2024voodooxp, tran2023voodoo, drobyshev2022megaportraits, Khakhulin2022ROME, goha}, potential misuse in the form of deepfakes is possible.
Developing strategies to detect such deepfakes is therefore of critical importance.
The field of passive forgery detection enables the identification of deepfakes without explicit watermarking \cite{agarwal2020detecting,cozzolino2021id,cozzolino2018forensictransfer,rossler2018faceforensics,rossler2019faceforensics++}.
However, generalized methods \cite{agarwal2020detecting, cozzolino2021id, cozzolino2018forensictransfer} have problems in reliably detecting fakes, and therefore cryptographical methods must be used in the future to verify the video's authenticity.

\end{document}